\pdfoutput=1
\documentclass[11pt]{article}
\usepackage[preprint]{acl}
\usepackage{times}
\usepackage{latexsym}
\usepackage{microtype}
\newif\ifacl\acltrue
\newcommand{\venuebibstyle}{acl_natbib}
\newenvironment{widetable}{\begin{table*}[t]}{\end{table*}}
\newenvironment{widefigure}{\begin{figure*}[t]}{\end{figure*}}
\newcommand{\floatwidth}{\textwidth}
\newcommand{\figonewidth}{0.52\textwidth}
\usepackage[utf8]{inputenc}
\usepackage[T1]{fontenc}
\usepackage{amsmath,amssymb,amsfonts,bm}
\usepackage{booktabs,multirow}
\usepackage{graphicx}
\usepackage{xcolor}
\usepackage{enumitem}
\usepackage{algorithm}
\usepackage{algpseudocode}
\usepackage{url}
\ifacl\hypersetup{colorlinks=true,linkcolor=blue!60!black,citecolor=blue!60!black,urlcolor=blue!60!black}\else
\usepackage[colorlinks=true,linkcolor=blue!60!black,citecolor=blue!60!black,urlcolor=blue!60!black]{hyperref}\fi
\usepackage{fvextra}
\usepackage{caption}
\newcommand{\figbox}[2]{\begin{center}\fbox{\parbox{0.92\linewidth}{\centering\vspace{#1}{\color{red!80!black}\small\textbf{FIGURE PLACEHOLDER.} #2}\vspace{#1}}}\end{center}}
\newcommand{\method}{LCC}
\newcommand{\methodlong}{Language-Consensus Control}
\newcommand{\ut}{u_t}
\DeclareMathOperator*{\argmin}{arg\,min}

\providecommand{\SamplingSingleDelta}{\pending{p7}}\providecommand{\SamplingSingleCI}{\pending{CI}}
\providecommand{\SamplingMedoidDelta}{\pending{p7}}\providecommand{\SamplingMedoidCI}{\pending{CI}}

\IfFileExists{tables/twobytwo_nums.tex}{

\providecommand{\SamplingSingleDelta}{}
\providecommand{\SamplingSingleCI}{}

\renewcommand{\SamplingSingleDelta}{-3.6}
\renewcommand{\SamplingSingleCI}{[-8.8, +1.6]}

\providecommand{\SamplingMedoidDelta}{}
\providecommand{\SamplingMedoidCI}{}

\renewcommand{\SamplingMedoidDelta}{+0.8}
\renewcommand{\SamplingMedoidCI}{[-4.4, +6.0]}

}{}

\providecommand{\AugMedoidDelta}{\pending{AugMedoidDelta}}
\providecommand{\AugMedoidCI}{\pending{AugMedoidCI}}
\providecommand{\UntouchedU}{\pending{UntouchedU}}
\providecommand{\UntouchedHist}{\pending{UntouchedHist}}
\providecommand{\UntouchedParaFrac}{\pending{UntouchedParaFrac}}

\providecommand{\OFTStdU}{\pending{OFTStdU}}

\providecommand{\AugU}{\pending{AugU}}
\providecommand{\UntouchedCanonFrac}{\pending{UntouchedCanonFrac}}
\providecommand{\AugCanonFrac}{\pending{AugCanonFrac}}
\providecommand{\ConsACanon}{\pending{ConsACanon}}
\providecommand{\ConsAPara}{\pending{ConsAPara}}
\providecommand{\ConsAHeldout}{\pending{ConsAHeldout}}
\providecommand{\ConsAWrong}{\pending{ConsAWrong}}
\providecommand{\ConsACost}{\pending{ConsACost}}
\providecommand{\ConsACostCI}{\pending{ConsACostCI}}

\providecommand{\ConsAShare}{\pending{ConsAShare}}

\providecommand{\ConsBCost}{\pending{ConsBCost}}
\providecommand{\ConsBCostCI}{\pending{ConsBCostCI}}
\providecommand{\HalfCost}{\pending{HalfCost}}
\providecommand{\HalfCostCI}{\pending{HalfCostCI}}
\providecommand{\ConsFirstCECanon}{\pending{ConsFirstCECanon}}
\providecommand{\ConsFirstCEPara}{\pending{ConsFirstCEPara}}
\providecommand{\ConsFirstKL}{\pending{ConsFirstKL}}
\providecommand{\ConsLastKL}{\pending{ConsLastKL}}
\providecommand{\ConsVsAugCanonDelta}{\pending{ConsVsAugCanonDelta}}
\providecommand{\ConsVsAugCanonCI}{\pending{ConsVsAugCanonCI}}
\providecommand{\ConsVsAugParaDelta}{\pending{ConsVsAugParaDelta}}
\providecommand{\ConsVsAugParaCI}{\pending{ConsVsAugParaCI}}
\providecommand{\ConsVsAugHeldoutDelta}{\pending{ConsVsAugHeldoutDelta}}
\providecommand{\ConsVsAugHeldoutCI}{\pending{ConsVsAugHeldoutCI}}

\IfFileExists{tables/aug_consensus_nums.tex}{

\providecommand{\AugMedoidDelta}{}
\providecommand{\AugMedoidCI}{}

\renewcommand{\AugMedoidDelta}{+3.2}
\renewcommand{\AugMedoidCI}{[-2.4, +8.8]}
}{}
\IfFileExists{tables/u_nums.tex}{\providecommand{\UntouchedU}{}\renewcommand{\UntouchedU}{0.281}

\providecommand{\UntouchedCanonFrac}{}\renewcommand{\UntouchedCanonFrac}{28}
\providecommand{\UntouchedParaFrac}{}\renewcommand{\UntouchedParaFrac}{72}
\providecommand{\UntouchedHist}{}\renewcommand{\UntouchedHist}{[0.28,0.25,0.26,0.21]}

\providecommand{\OFTStdU}{}\renewcommand{\OFTStdU}{0.024}

\providecommand{\AugU}{}\renewcommand{\AugU}{0.204}

\providecommand{\AugCanonFrac}{}\renewcommand{\AugCanonFrac}{50}

}{}
\IfFileExists{tables/cons_nums.tex}{\providecommand{\ConsACanon}{}
\renewcommand{\ConsACanon}{84.4}
\providecommand{\ConsAPara}{}
\renewcommand{\ConsAPara}{81.6}
\providecommand{\ConsAHeldout}{}
\renewcommand{\ConsAHeldout}{81.2}
\providecommand{\ConsAWrong}{}
\renewcommand{\ConsAWrong}{4.8}
\providecommand{\ConsACost}{}
\renewcommand{\ConsACost}{-2.8}
\providecommand{\ConsACostCI}{}
\renewcommand{\ConsACostCI}{[-8.0, +2.4]}

\providecommand{\ConsAShare}{}
\renewcommand{\ConsAShare}{94}

\providecommand{\ConsBCost}{}
\renewcommand{\ConsBCost}{+1.2}
\providecommand{\ConsBCostCI}{}
\renewcommand{\ConsBCostCI}{[-4.4, +7.2]}

\providecommand{\HalfCost}{}
\renewcommand{\HalfCost}{-4.0}
\providecommand{\HalfCostCI}{}
\renewcommand{\HalfCostCI}{[-10.0, +2.0]}
}{}
\IfFileExists{tables/cons_train_nums.tex}{\providecommand{\ConsFirstCECanon}{}
\renewcommand{\ConsFirstCECanon}{0.0002}
\providecommand{\ConsFirstCEPara}{}
\renewcommand{\ConsFirstCEPara}{2.68}
\providecommand{\ConsFirstKL}{}
\renewcommand{\ConsFirstKL}{6.62}
\providecommand{\ConsLastKL}{}
\renewcommand{\ConsLastKL}{0.06}
}{}
\IfFileExists{tables/cons_vs_aug_nums.tex}{\providecommand{\ConsVsAugCanonDelta}{}\renewcommand{\ConsVsAugCanonDelta}{+2.8}
\providecommand{\ConsVsAugCanonCI}{}\renewcommand{\ConsVsAugCanonCI}{[-2.8, +8.4]}

\providecommand{\ConsVsAugParaDelta}{}\renewcommand{\ConsVsAugParaDelta}{-2.0}
\providecommand{\ConsVsAugParaCI}{}\renewcommand{\ConsVsAugParaCI}{[-7.2, +3.2]}

\providecommand{\ConsVsAugHeldoutDelta}{}\renewcommand{\ConsVsAugHeldoutDelta}{-2.4}
\providecommand{\ConsVsAugHeldoutCI}{}\renewcommand{\ConsVsAugHeldoutCI}{[-8.0, +3.2]}

}{}
\providecommand{\UntouchedHeldoutCost}{\pending{x}}\providecommand{\UntouchedHeldoutCostCI}{\pending{x}}\providecommand{\CtrlHeldoutCost}{\pending{x}}\providecommand{\CtrlHeldoutCostCI}{\pending{x}}
\IfFileExists{tables/p19_nums.tex}{\providecommand{\AugVsCtrlHeldDelta}{}\renewcommand{\AugVsCtrlHeldDelta}{+27.2}
\providecommand{\AugVsCtrlHeldCI}{}\renewcommand{\AugVsCtrlHeldCI}{[+20.8, +34.0]}

\providecommand{\AugVsCtrlParaDelta}{}\renewcommand{\AugVsCtrlParaDelta}{+32.8}
\providecommand{\AugVsCtrlParaCI}{}\renewcommand{\AugVsCtrlParaCI}{[+25.6, +40.0]}

\providecommand{\AugVsCtrlCanonDelta}{}\renewcommand{\AugVsCtrlCanonDelta}{-1.2}
\providecommand{\AugVsCtrlCanonCI}{}\renewcommand{\AugVsCtrlCanonCI}{[-7.2, +4.8]}

\providecommand{\UntouchedHeldVsParaDelta}{}\renewcommand{\UntouchedHeldVsParaDelta}{+10.0}
\providecommand{\UntouchedHeldVsParaCI}{}\renewcommand{\UntouchedHeldVsParaCI}{[+2.8, +17.6]}

}{}
\IfFileExists{tables/text_nums.tex}{\providecommand{\BaseMedoidMag}{}\renewcommand{\BaseMedoidMag}{16.8}%
\providecommand{\BaseMedoidRate}{}\renewcommand{\BaseMedoidRate}{68.8}%
\providecommand{\BaseMedoidPairedCI}{}\renewcommand{\BaseMedoidPairedCI}{[-22.8, -10.8]}%
\providecommand{\BaseMedoidDiscPct}{}\renewcommand{\BaseMedoidDiscPct}{28}%
\providecommand{\OneOfKEpisodeMag}{}\renewcommand{\OneOfKEpisodeMag}{27.2}%
\providecommand{\OneOfKReplanMag}{}\renewcommand{\OneOfKReplanMag}{19.2}%
\providecommand{\BaseReplDiff}{}\renewcommand{\BaseReplDiff}{2.0}%
\providecommand{\BaseHeadroom}{}\renewcommand{\BaseHeadroom}{14.4}%
\providecommand{\BaseLongMedoidDelta}{}\renewcommand{\BaseLongMedoidDelta}{-7.0}%
\providecommand{\BaseLongMedoidCI}{}\renewcommand{\BaseLongMedoidCI}{[-18.0, +4.0]}%
\providecommand{\BaseLongMedoidN}{}\renewcommand{\BaseLongMedoidN}{100}%
\providecommand{\BaseKEightDelta}{}\renewcommand{\BaseKEightDelta}{-10.0}%
\providecommand{\BaseKEightCI}{}\renewcommand{\BaseKEightCI}{[-21.0, +1.0]}%
\providecommand{\BaseKEightN}{}\renewcommand{\BaseKEightN}{100}%
\providecommand{\BaseLongCanonProbeRef}{}\renewcommand{\BaseLongCanonProbeRef}{55.0}%
\providecommand{\BaseLongCanonConsRef}{}\renewcommand{\BaseLongCanonConsRef}{50.0}%
\providecommand{\BaseLongReplDiff}{}\renewcommand{\BaseLongReplDiff}{5.0}%
\providecommand{\OftShiftMedoidRate}{}\renewcommand{\OftShiftMedoidRate}{90.0}%
\providecommand{\OftShiftCanonRate}{}\renewcommand{\OftShiftCanonRate}{88.6}%
\providecommand{\OftShiftMedoidDelta}{}\renewcommand{\OftShiftMedoidDelta}{+1.4}%
\providecommand{\OftShiftMedoidCI}{}\renewcommand{\OftShiftMedoidCI}{[-0.6, +3.4]}%
\providecommand{\OftShiftMedoidPHolm}{}\renewcommand{\OftShiftMedoidPHolm}{0.74}%
\providecommand{\OftShiftMedoidN}{}\renewcommand{\OftShiftMedoidN}{500}%
\providecommand{\OftShiftHeadroom}{}\renewcommand{\OftShiftHeadroom}{11.4}%
\providecommand{\OftStdCanonRate}{}\renewcommand{\OftStdCanonRate}{97.2}%
\providecommand{\OftStdMaxAbsDelta}{}\renewcommand{\OftStdMaxAbsDelta}{1.6}%
\providecommand{\OftReplDiff}{}\renewcommand{\OftReplDiff}{0.8}%
\providecommand{\OftGoalMedoidMag}{}\renewcommand{\OftGoalMedoidMag}{4.0}%
\providecommand{\HarnessDiscPct}{}\renewcommand{\HarnessDiscPct}{4}%
\providecommand{\MdePts}{}\renewcommand{\MdePts}{2.6}%
\providecommand{\OftSpatialWrongPct}{}\renewcommand{\OftSpatialWrongPct}{74}%
\providecommand{\OftObjectWrongPct}{}\renewcommand{\OftObjectWrongPct}{96}%
\providecommand{\LatencyKFourRatio}{}\renewcommand{\LatencyKFourRatio}{2.98}%
\providecommand{\GoalAugCanonDelta}{}\renewcommand{\GoalAugCanonDelta}{+2.4}%
\providecommand{\GoalAugCanonCI}{}\renewcommand{\GoalAugCanonCI}{[-3.6, +8.4]}%
}{}
\IfFileExists{tables/seed_nums.tex}{

\providecommand{\AugSeedBCost}{}
\renewcommand{\AugSeedBCost}{-2.0}

\providecommand{\AugSeedBHeldoutCost}{}
\renewcommand{\AugSeedBHeldoutCost}{+0.8}

\providecommand{\CtrlSeedBCost}{}
\renewcommand{\CtrlSeedBCost}{-36.8}

\providecommand{\CtrlSeedBHeldoutCost}{}
\renewcommand{\CtrlSeedBHeldoutCost}{-27.6}

}{}
\IfFileExists{tables/seed_cmp_nums.tex}{\providecommand{\SeedBAugVsCtrlParaDelta}{}\renewcommand{\SeedBAugVsCtrlParaDelta}{+38.0}
\providecommand{\SeedBAugVsCtrlParaCI}{}\renewcommand{\SeedBAugVsCtrlParaCI}{[+30.8, +44.8]}

\providecommand{\SeedBAugVsCtrlHeldDelta}{}\renewcommand{\SeedBAugVsCtrlHeldDelta}{+31.6}
\providecommand{\SeedBAugVsCtrlHeldCI}{}\renewcommand{\SeedBAugVsCtrlHeldCI}{[+25.2, +38.4]}

\providecommand{\SeedBVsAAugParaDelta}{}\renewcommand{\SeedBVsAAugParaDelta}{-2.4}
\providecommand{\SeedBVsAAugParaCI}{}\renewcommand{\SeedBVsAAugParaCI}{[-8.0, +3.2]}

\providecommand{\SeedBAugCanonDropDelta}{}\renewcommand{\SeedBAugCanonDropDelta}{-2.4}

\providecommand{\SeedBCtrlCanonDropDelta}{}\renewcommand{\SeedBCtrlCanonDropDelta}{-5.6}
\providecommand{\SeedBCtrlCanonDropCI}{}\renewcommand{\SeedBCtrlCanonDropCI}{[-11.2, -0.4]}

}{}
\IfFileExists{tables/conf_nums.tex}{\providecommand{\ParaConfVsOneDelta}{}\renewcommand{\ParaConfVsOneDelta}{+10.0}
\providecommand{\ParaConfVsOneCI}{}\renewcommand{\ParaConfVsOneCI}{[+3.2, +16.8]}

\providecommand{\ParaConfVsOneRate}{}\renewcommand{\ParaConfVsOneRate}{57.6}
\providecommand{\ParaConfVsOneClCI}{}\renewcommand{\ParaConfVsOneClCI}{[-1.6, +22.0]}

\providecommand{\ParaConfVsCanonDelta}{}\renewcommand{\ParaConfVsCanonDelta}{-28.0}
\providecommand{\ParaConfVsCanonCI}{}\renewcommand{\ParaConfVsCanonCI}{[-34.4, -21.2]}

\providecommand{\ConfVsCanonDelta}{}\renewcommand{\ConfVsCanonDelta}{-13.0}
\providecommand{\ConfVsCanonCI}{}\renewcommand{\ConfVsCanonCI}{[-23.0, -3.0]}

\providecommand{\ConfVsCanonN}{}\renewcommand{\ConfVsCanonN}{100}

\providecommand{\ConfCanonFrac}{}\renewcommand{\ConfCanonFrac}{37}
\providecommand{\ConfMedoidCanonFrac}{}\renewcommand{\ConfMedoidCanonFrac}{28}
}{}
\providecommand{\UntouchedCanon}{\pending{UntouchedCanon}}
\providecommand{\UntouchedPara}{\pending{UntouchedPara}}
\providecommand{\UntouchedCost}{\pending{UntouchedCost}}

\providecommand{\AugCanon}{\pending{AugCanon}}

\providecommand{\AugHeldout}{\pending{AugHeldout}}
\providecommand{\AugWrong}{\pending{AugWrong}}
\providecommand{\AugCost}{\pending{AugCost}}
\providecommand{\AugCostCI}{\pending{AugCostCI}}
\providecommand{\AugHeldoutCost}{\pending{AugHeldoutCost}}
\providecommand{\AugHeldoutCostCI}{\pending{AugHeldoutCostCI}}
\providecommand{\AugShare}{\pending{AugShare}}
\providecommand{\CtrlCanon}{\pending{CtrlCanon}}

\providecommand{\CtrlCost}{\pending{CtrlCost}}
\providecommand{\CtrlCostCI}{\pending{CtrlCostCI}}
\providecommand{\UntouchedEasyDelta}{\pending{UntouchedEasyDelta}}

\providecommand{\UntouchedHardDelta}{\pending{UntouchedHardDelta}}

\providecommand{\UntouchedAdversarialDelta}{\pending{UntouchedAdversarialDelta}}

\providecommand{\AugEasyDelta}{\pending{AugEasyDelta}}

\providecommand{\AugHardDelta}{\pending{AugHardDelta}}

\providecommand{\AugAdversarialDelta}{\pending{AugAdversarialDelta}}
\providecommand{\AugAdversarialCI}{\pending{AugAdversarialCI}}

\providecommand{\GoalUntouchedCost}{\pending{GoalUntouchedCost}}

\providecommand{\GoalAugCost}{\pending{GoalAugCost}}
\providecommand{\GoalAugCostCI}{\pending{GoalAugCostCI}}

\providecommand{\GoalCtrlCost}{\pending{GoalCtrlCost}}
\providecommand{\GoalCtrlCostCI}{\pending{GoalCtrlCostCI}}

\IfFileExists{tables/aug_nums.tex}{\providecommand{\UntouchedCanon}{}
\renewcommand{\UntouchedCanon}{85.6}
\providecommand{\UntouchedPara}{}
\renewcommand{\UntouchedPara}{47.6}

\providecommand{\UntouchedWrong}{}
\renewcommand{\UntouchedWrong}{2.4}
\providecommand{\UntouchedCost}{}
\renewcommand{\UntouchedCost}{-38.0}

\providecommand{\UntouchedHeldoutCost}{}
\renewcommand{\UntouchedHeldoutCost}{-28.0}
\providecommand{\UntouchedHeldoutCostCI}{}
\renewcommand{\UntouchedHeldoutCostCI}{[-34.4, -21.6]}

\providecommand{\AugCanon}{}
\renewcommand{\AugCanon}{81.6}

\providecommand{\AugHeldout}{}
\renewcommand{\AugHeldout}{83.6}
\providecommand{\AugWrong}{}
\renewcommand{\AugWrong}{2.8}
\providecommand{\AugCost}{}
\renewcommand{\AugCost}{+2.0}
\providecommand{\AugCostCI}{}
\renewcommand{\AugCostCI}{[-4.0, +8.0]}
\providecommand{\AugHeldoutCost}{}
\renewcommand{\AugHeldoutCost}{+2.0}
\providecommand{\AugHeldoutCostCI}{}
\renewcommand{\AugHeldoutCostCI}{[-3.2, +7.6]}
\providecommand{\AugShare}{}
\renewcommand{\AugShare}{97}
\providecommand{\CtrlCanon}{}
\renewcommand{\CtrlCanon}{82.8}

\providecommand{\CtrlCost}{}
\renewcommand{\CtrlCost}{-32.0}
\providecommand{\CtrlCostCI}{}
\renewcommand{\CtrlCostCI}{[-38.8, -25.2]}
\providecommand{\CtrlHeldoutCost}{}
\renewcommand{\CtrlHeldoutCost}{-26.4}
\providecommand{\CtrlHeldoutCostCI}{}
\renewcommand{\CtrlHeldoutCostCI}{[-33.2, -19.2]}

\providecommand{\HalfCost}{}
\renewcommand{\HalfCost}{-4.0}
\providecommand{\HalfCostCI}{}
\renewcommand{\HalfCostCI}{[-10.0, +2.0]}
}{}
\IfFileExists{tables/goal_nums.tex}{

\providecommand{\GoalUntouchedCost}{}
\renewcommand{\GoalUntouchedCost}{-36.8}

\providecommand{\GoalAugWrong}{}
\renewcommand{\GoalAugWrong}{0.0}
\providecommand{\GoalAugCost}{}
\renewcommand{\GoalAugCost}{-4.0}
\providecommand{\GoalAugCostCI}{}
\renewcommand{\GoalAugCostCI}{[-10.0, +2.0]}

\providecommand{\GoalCtrlCost}{}
\renewcommand{\GoalCtrlCost}{-43.6}
\providecommand{\GoalCtrlCostCI}{}
\renewcommand{\GoalCtrlCostCI}{[-50.4, -36.4]}
}{}
\providecommand{\OftGoalUntouchedCost}{\pending{OftGoalUntouchedCost}}\providecommand{\OftGoalAugCost}{\pending{OftGoalAugCost}}\providecommand{\OftGoalAugCostCI}{\pending{OftGoalAugCostCI}}\providecommand{\OftGoalCtrlCost}{\pending{OftGoalCtrlCost}}\providecommand{\OftGoalUntouchedHeldoutCost}{\pending{x}}\providecommand{\OftGoalAugHeldoutCost}{\pending{x}}\providecommand{\OftGoalCtrlHeldoutCost}{\pending{x}}\providecommand{\OftGoalAugShare}{\pending{x}}
\IfFileExists{tables/oft_goal_aug_nums.tex}{

\providecommand{\OftGoalUntouchedCost}{}
\renewcommand{\OftGoalUntouchedCost}{-10.4}

\providecommand{\OftGoalUntouchedHeldoutCost}{}
\renewcommand{\OftGoalUntouchedHeldoutCost}{-8.0}

\providecommand{\OftGoalAugCost}{}
\renewcommand{\OftGoalAugCost}{+0.0}
\providecommand{\OftGoalAugCostCI}{}
\renewcommand{\OftGoalAugCostCI}{[-2.4, +2.0]}
\providecommand{\OftGoalAugHeldoutCost}{}
\renewcommand{\OftGoalAugHeldoutCost}{-2.0}

\providecommand{\OftGoalAugShare}{}
\renewcommand{\OftGoalAugShare}{100}

\providecommand{\OftGoalCtrlCost}{}
\renewcommand{\OftGoalCtrlCost}{-10.0}

\providecommand{\OftGoalCtrlHeldoutCost}{}
\renewcommand{\OftGoalCtrlHeldoutCost}{-7.2}

}{}
\providecommand{\OftGoalMedoidDelta}{\pending{x}}\providecommand{\OftGoalMedoidCI}{\pending{x}}\providecommand{\OftGoalParaMedoidVsOneDelta}{\pending{x}}\providecommand{\OftGoalParaMedoidVsOneCI}{\pending{x}}\providecommand{\OftGoalParaMedoidVsOneP}{\pending{x}}\providecommand{\OftGoalParaMedoidDelta}{\pending{x}}\providecommand{\OftGoalAugCanonDropDelta}{\pending{x}}\providecommand{\OftGoalAugCanonDropCI}{\pending{x}}\providecommand{\OftGoalCtrlCanonDropDelta}{\pending{x}}\providecommand{\OftGoalCtrlCanonDropCI}{\pending{x}}\providecommand{\OftGoalAugVsCtrlHeldDelta}{\pending{x}}\providecommand{\OftGoalAugVsCtrlHeldCI}{\pending{x}}\providecommand{\OftGoalAugVsCtrlParaDelta}{\pending{x}}\providecommand{\OftGoalAugVsCtrlParaCI}{\pending{x}}\providecommand{\TrainRuns}{\pending{x}}
\IfFileExists{tables/oft_goal_consensus_nums.tex}{\providecommand{\OftGoalMedoidDelta}{}\renewcommand{\OftGoalMedoidDelta}{-4.0}
\providecommand{\OftGoalMedoidCI}{}\renewcommand{\OftGoalMedoidCI}{[-6.4, -1.6]}

\providecommand{\OftGoalParaMedoidDelta}{}\renewcommand{\OftGoalParaMedoidDelta}{-6.8}

\providecommand{\OftGoalParaMedoidVsOneDelta}{}\renewcommand{\OftGoalParaMedoidVsOneDelta}{+3.6}
\providecommand{\OftGoalParaMedoidVsOneCI}{}\renewcommand{\OftGoalParaMedoidVsOneCI}{[+0.4, +7.2]}
\providecommand{\OftGoalParaMedoidVsOneP}{}\renewcommand{\OftGoalParaMedoidVsOneP}{0.06}
}{}
\IfFileExists{tables/oft_goal_contrasts_nums.tex}{\providecommand{\OftGoalAugCanonDropDelta}{}\renewcommand{\OftGoalAugCanonDropDelta}{-3.2}
\providecommand{\OftGoalAugCanonDropCI}{}\renewcommand{\OftGoalAugCanonDropCI}{[-5.6, -1.2]}

\providecommand{\OftGoalCtrlCanonDropDelta}{}\renewcommand{\OftGoalCtrlCanonDropDelta}{-2.8}
\providecommand{\OftGoalCtrlCanonDropCI}{}\renewcommand{\OftGoalCtrlCanonDropCI}{[-5.2, -1.2]}

\providecommand{\OftGoalAugVsCtrlHeldDelta}{}\renewcommand{\OftGoalAugVsCtrlHeldDelta}{+4.8}
\providecommand{\OftGoalAugVsCtrlHeldCI}{}\renewcommand{\OftGoalAugVsCtrlHeldCI}{[+0.8, +8.8]}

\providecommand{\OftGoalAugVsCtrlParaDelta}{}\renewcommand{\OftGoalAugVsCtrlParaDelta}{+9.6}
\providecommand{\OftGoalAugVsCtrlParaCI}{}\renewcommand{\OftGoalAugVsCtrlParaCI}{[+5.6, +13.6]}

}{}

\providecommand{\OftObjectShare}{\pending{OftObjectShare}}
\providecommand{\OftGoalShare}{\pending{OftGoalShare}}

\providecommand{\BaseSpatialShare}{\pending{BaseSpatialShare}}

\providecommand{\BaseGoalShare}{\pending{BaseGoalShare}}

\providecommand{\OftSpatialParaDelta}{\pending{OftSpatialParaDelta}}
\providecommand{\OftObjectParaDelta}{\pending{OftObjectParaDelta}}
\providecommand{\OftGoalParaDelta}{\pending{OftGoalParaDelta}}
\providecommand{\OftLongParaDelta}{\pending{OftLongParaDelta}}
\providecommand{\BaseSpatialParaDelta}{\pending{BaseSpatialParaDelta}}
\providecommand{\BaseObjectParaDelta}{\pending{BaseObjectParaDelta}}
\providecommand{\BaseGoalParaDelta}{\pending{BaseGoalParaDelta}}

\providecommand{\OftGoalParaCI}{\pending{OftGoalParaCI}}

\providecommand{\SamplingVsParaMedoidDelta}{\pending{SamplingVsParaMedoidDelta}}
\providecommand{\SamplingVsParaMedoidCI}{\pending{SamplingVsParaMedoidCI}}
\providecommand{\EvalGPUHours}{\pending{EvalGPUHours}}
\providecommand{\EvalProcessHours}{\pending{EvalProcessHours}}
\providecommand{\TotalEpisodes}{\pending{TotalEpisodes}}
\providecommand{\TrainHours}{\pending{TrainHours}}
\providecommand{\PiSpatialCanon}{\pending{PiSpatialCanon}}\providecommand{\PiObjectCanon}{\pending{PiObjectCanon}}\providecommand{\PiGoalCanon}{\pending{PiGoalCanon}}\providecommand{\PiLongCanon}{\pending{PiLongCanon}}
\providecommand{\PiSpatialShare}{\pending{PiSpatialShare}}\providecommand{\PiObjectShare}{\pending{PiObjectShare}}\providecommand{\PiGoalShare}{\pending{PiGoalShare}}\providecommand{\PiLongShare}{\pending{PiLongShare}}
\providecommand{\PiObjectParaDelta}{\pending{PiObjectParaDelta}}\providecommand{\PiGoalParaDelta}{\pending{PiGoalParaDelta}}\providecommand{\PiGoalParaCI}{\pending{PiGoalParaCI}}
\IfFileExists{tables/pi05_nums.tex}{\providecommand{\PiLongCanon}{}\renewcommand{\PiLongCanon}{95.0}

\providecommand{\PiLongShare}{}\renewcommand{\PiLongShare}{39}
\providecommand{\PiGoalCanon}{}\renewcommand{\PiGoalCanon}{97.2}

\providecommand{\PiGoalShare}{}\renewcommand{\PiGoalShare}{100}
\providecommand{\PiObjectCanon}{}\renewcommand{\PiObjectCanon}{98.0}

\providecommand{\PiObjectShare}{}\renewcommand{\PiObjectShare}{67}
\providecommand{\PiSpatialCanon}{}\renewcommand{\PiSpatialCanon}{98.8}

\providecommand{\PiSpatialShare}{}\renewcommand{\PiSpatialShare}{27}
}{}
\IfFileExists{tables/reliance_nums.tex}{

\providecommand{\OftSpatialParaDelta}{}\renewcommand{\OftSpatialParaDelta}{-1.0}

\providecommand{\OftObjectShare}{}\renewcommand{\OftObjectShare}{2}

\providecommand{\OftObjectParaDelta}{}\renewcommand{\OftObjectParaDelta}{-1.6}

\providecommand{\OftGoalShare}{}\renewcommand{\OftGoalShare}{100}

\providecommand{\OftGoalLost}{}\renewcommand{\OftGoalLost}{10}
\providecommand{\OftGoalParaDelta}{}\renewcommand{\OftGoalParaDelta}{-10.4}
\providecommand{\OftGoalParaCI}{}\renewcommand{\OftGoalParaCI}{[-14.4, -6.8]}

\providecommand{\OftLongParaDelta}{}\renewcommand{\OftLongParaDelta}{-3.0}

\providecommand{\BaseSpatialShare}{}\renewcommand{\BaseSpatialShare}{97}

\providecommand{\BaseSpatialLost}{}\renewcommand{\BaseSpatialLost}{46}
\providecommand{\BaseSpatialParaDelta}{}\renewcommand{\BaseSpatialParaDelta}{-38.0}

\providecommand{\BaseObjectParaDelta}{}\renewcommand{\BaseObjectParaDelta}{-15.6}

\providecommand{\BaseGoalShare}{}\renewcommand{\BaseGoalShare}{100}

\providecommand{\BaseGoalLost}{}\renewcommand{\BaseGoalLost}{48}
\providecommand{\BaseGoalParaDelta}{}\renewcommand{\BaseGoalParaDelta}{-36.8}

\providecommand{\PiSpatialShare}{}\renewcommand{\PiSpatialShare}{27}

\providecommand{\PiObjectShare}{}\renewcommand{\PiObjectShare}{67}

\providecommand{\PiObjectParaDelta}{}\renewcommand{\PiObjectParaDelta}{+1.0}

\providecommand{\PiGoalShare}{}\renewcommand{\PiGoalShare}{100}

\providecommand{\PiGoalLost}{}\renewcommand{\PiGoalLost}{7}
\providecommand{\PiGoalParaDelta}{}\renewcommand{\PiGoalParaDelta}{-7.2}
\providecommand{\PiGoalParaCI}{}\renewcommand{\PiGoalParaCI}{[-11.2, -3.2]}

\providecommand{\PiLongShare}{}\renewcommand{\PiLongShare}{39}

}{}
\providecommand{\OftGoalLost}{\pending{OftGoalLost}}\providecommand{\PiGoalLost}{\pending{PiGoalLost}}\providecommand{\BaseSpatialLost}{\pending{BaseSpatialLost}}\providecommand{\BaseGoalLost}{\pending{BaseGoalLost}}
\IfFileExists{tables/reliance_task_nums.tex}{

}{}
\IfFileExists{tables/sampling_vs_para_nums.tex}{\providecommand{\SamplingVsParaMedoidDelta}{}\renewcommand{\SamplingVsParaMedoidDelta}{+17.6}
\providecommand{\SamplingVsParaMedoidCI}{}\renewcommand{\SamplingVsParaMedoidCI}{[+11.2, +24.0]}

}{}
\IfFileExists{tables/compute_nums.tex}{\providecommand{\TotalEpisodes}{}\renewcommand{\TotalEpisodes}{40{,}440}
\providecommand{\EvalProcessHours}{}\renewcommand{\EvalProcessHours}{627}
\providecommand{\EvalGPUHours}{}\renewcommand{\EvalGPUHours}{232}
\providecommand{\TrainHours}{}\renewcommand{\TrainHours}{16.6}
\providecommand{\TrainRuns}{}\renewcommand{\TrainRuns}{14}
}{}
\providecommand{\AugAdversarialPHolm}{\pending{AugAdversarialPHolm}}
\IfFileExists{tables/variant_nums.tex}{

\providecommand{\AugEasyDelta}{}\renewcommand{\AugEasyDelta}{+0.8}

\providecommand{\AugHardDelta}{}\renewcommand{\AugHardDelta}{-2.4}

\providecommand{\AugAdversarialDelta}{}\renewcommand{\AugAdversarialDelta}{-6.8}
\providecommand{\AugAdversarialCI}{}\renewcommand{\AugAdversarialCI}{[-13.2, -0.4]}

\providecommand{\AugAdversarialPHolm}{}\renewcommand{\AugAdversarialPHolm}{0.249}

\providecommand{\UntouchedEasyDelta}{}\renewcommand{\UntouchedEasyDelta}{-14.8}

\providecommand{\UntouchedHardDelta}{}\renewcommand{\UntouchedHardDelta}{-48.0}

\providecommand{\UntouchedAdversarialDelta}{}\renewcommand{\UntouchedAdversarialDelta}{-28.8}

}{}
\IfFileExists{tables/paraphrase_only_nums.tex}{

}{}
\providecommand{\CanonRetN}{\pending{CanonRetN}}\providecommand{\CanonRetTrigram}{\pending{CanonRetTrigram}}\providecommand{\CanonRetTrigramWorst}{\pending{CanonRetTrigramWorst}}
\IfFileExists{tables/canon_retrieval_nums.tex}{\providecommand{\CanonRetN}{}\renewcommand{\CanonRetN}{1{,}398}
\providecommand{\CanonRetTrigram}{}\renewcommand{\CanonRetTrigram}{95.5}

\providecommand{\CanonRetTrigramWorst}{}\renewcommand{\CanonRetTrigramWorst}{75.7}

}{}
\providecommand{\HumanN}{\pending{HumanN}}\providecommand{\HumanDiff}{\pending{HumanDiff}}\providecommand{\HumanDiffPct}{\pending{HumanDiffPct}}\providecommand{\AugLlmHuman}{\pending{AugLlmHuman}}\providecommand{\CtrlLlmHuman}{\pending{CtrlLlmHuman}}
\providecommand{\MedoidEquivTasks}{\pending{MedoidEquivTasks}}\providecommand{\MedoidEquivTasksCI}{\pending{MedoidEquivTasksCI}}\providecommand{\MedoidEquivTasksN}{\pending{MedoidEquivTasksN}}\providecommand{\MedoidOtherTasks}{\pending{MedoidOtherTasks}}\providecommand{\MedoidOtherTasksCI}{\pending{MedoidOtherTasksCI}}\providecommand{\AugCanonDropDelta}{\pending{AugCanonDropDelta}}\providecommand{\AugCanonDropCI}{\pending{AugCanonDropCI}}\providecommand{\CtrlCanonDropDelta}{\pending{CtrlCanonDropDelta}}\providecommand{\CtrlCanonDropCI}{\pending{CtrlCanonDropCI}}
\IfFileExists{tables/human_equiv_nums.tex}{\providecommand{\HumanN}{}\renewcommand{\HumanN}{240}

\providecommand{\HumanDiff}{}\renewcommand{\HumanDiff}{78}
\providecommand{\HumanDiffPct}{}\renewcommand{\HumanDiffPct}{32}
\providecommand{\MedoidEquivTasks}{}\renewcommand{\MedoidEquivTasks}{-28.0}
\providecommand{\MedoidEquivTasksCI}{}\renewcommand{\MedoidEquivTasksCI}{[-42.0, -14.0]}
\providecommand{\MedoidEquivTasksN}{}\renewcommand{\MedoidEquivTasksN}{2}
\providecommand{\MedoidOtherTasks}{}\renewcommand{\MedoidOtherTasks}{-14.0}
\providecommand{\MedoidOtherTasksCI}{}\renewcommand{\MedoidOtherTasksCI}{[-21.0, -7.0]}

\providecommand{\OftGoalHumanDiff}{}\renewcommand{\OftGoalHumanDiff}{-17.3}

\providecommand{\OftGoalHuman}{}\renewcommand{\OftGoalHuman}{-8.6}

\providecommand{\BaseGoalHumanDiff}{}\renewcommand{\BaseGoalHumanDiff}{-55.8}

\providecommand{\BaseGoalHuman}{}\renewcommand{\BaseGoalHuman}{-31.8}

\providecommand{\AugLlmHumanDiff}{}\renewcommand{\AugLlmHumanDiff}{-1.9}

\providecommand{\AugLlmHuman}{}\renewcommand{\AugLlmHuman}{+4.9}

\providecommand{\CtrlLlmHumanDiff}{}\renewcommand{\CtrlLlmHumanDiff}{-40.2}

\providecommand{\CtrlLlmHuman}{}\renewcommand{\CtrlLlmHuman}{-25.9}

}{}
\IfFileExists{tables/canon_drop_nums.tex}{\providecommand{\AugCanonDropDelta}{}\renewcommand{\AugCanonDropDelta}{-4.0}
\providecommand{\AugCanonDropCI}{}\renewcommand{\AugCanonDropCI}{[-9.6, +2.0]}

\providecommand{\CtrlCanonDropDelta}{}\renewcommand{\CtrlCanonDropDelta}{-2.8}
\providecommand{\CtrlCanonDropCI}{}\renewcommand{\CtrlCanonDropCI}{[-8.0, +2.4]}

}{}
\providecommand{\ParaMedoidVsOneDelta}{\pending{ParaMedoidVsOneDelta}}\providecommand{\ParaMedoidVsOneCI}{\pending{ParaMedoidVsOneCI}}\providecommand{\ParaMedoidVsOneP}{\pending{ParaMedoidVsOneP}}\providecommand{\ParaMedoidVsCanonDelta}{\pending{ParaMedoidVsCanonDelta}}\providecommand{\ParaMedoidVsCanonCI}{\pending{ParaMedoidVsCanonCI}}\providecommand{\ParaMedoidVsCanonMedoidDelta}{\pending{ParaMedoidVsCanonMedoidDelta}}\providecommand{\ParaMedoidRate}{\pending{ParaMedoidRate}}
\IfFileExists{tables/para_medoid_nums.tex}{\providecommand{\ParaMedoidVsOneDelta}{}\renewcommand{\ParaMedoidVsOneDelta}{+5.6}
\providecommand{\ParaMedoidVsOneCI}{}\renewcommand{\ParaMedoidVsOneCI}{[-0.8, +12.0]}
\providecommand{\ParaMedoidVsOneP}{}\renewcommand{\ParaMedoidVsOneP}{0.12}
\providecommand{\ParaMedoidVsCanonDelta}{}\renewcommand{\ParaMedoidVsCanonDelta}{-32.4}
\providecommand{\ParaMedoidVsCanonCI}{}\renewcommand{\ParaMedoidVsCanonCI}{[-39.2, -25.6]}

\providecommand{\ParaMedoidVsCanonMedoidDelta}{}\renewcommand{\ParaMedoidVsCanonMedoidDelta}{-15.6}

}{}
\IfFileExists{tables/para_medoid_rate_nums.tex}{\providecommand{\ParaMedoidRate}{}

\renewcommand{\ParaMedoidRate}{53.2}

}{}
\providecommand{\ParaSampMedoidVsSingleDelta}{\pending{x}}\providecommand{\ParaSampMedoidVsSingleCI}{\pending{x}}\providecommand{\ParaSampMedoidVsCanonDelta}{\pending{x}}\providecommand{\ParaSampMedoidVsCanonCI}{\pending{x}}\providecommand{\ParaSampMedoidVsGreedyDelta}{\pending{x}}\providecommand{\ParaSampMedoidVsGreedyCI}{\pending{x}}\providecommand{\ParaSampClCI}{\pending{x}}\providecommand{\ParaSampMedoidRate}{\pending{x}}
\IfFileExists{tables/para_sampling_nums.tex}{\providecommand{\ParaSampMedoidVsGreedyDelta}{}\renewcommand{\ParaSampMedoidVsGreedyDelta}{+10.4}
\providecommand{\ParaSampMedoidVsGreedyCI}{}\renewcommand{\ParaSampMedoidVsGreedyCI}{[+4.8, +16.0]}

\providecommand{\ParaSampMedoidVsSingleDelta}{}\renewcommand{\ParaSampMedoidVsSingleDelta}{+6.4}
\providecommand{\ParaSampMedoidVsSingleCI}{}\renewcommand{\ParaSampMedoidVsSingleCI}{[+1.2, +11.6]}

\providecommand{\ParaSampMedoidVsCanonDelta}{}\renewcommand{\ParaSampMedoidVsCanonDelta}{-27.6}
\providecommand{\ParaSampMedoidVsCanonCI}{}\renewcommand{\ParaSampMedoidVsCanonCI}{[-34.8, -20.8]}

}{}
\IfFileExists{tables/para_sampling_rate_nums.tex}{\providecommand{\ParaSampMedoidRate}{}\renewcommand{\ParaSampMedoidRate}{58.0}
}{}
\providecommand{\ParaMedoidClCI}{\pending{ParaMedoidClCI}}\providecommand{\BaseTrigramRho}{\pending{BaseTrigramRho}}\providecommand{\BaseEmbedRho}{\pending{BaseEmbedRho}}\providecommand{\BaseParaN}{\pending{BaseParaN}}\providecommand{\BaseHumanRho}{\pending{BaseHumanRho}}
\IfFileExists{tables/clustered_nums.tex}{

\providecommand{\ParaMedoidClCI}{}\renewcommand{\ParaMedoidClCI}{[-4.0, +16.0]}
\providecommand{\ParaSampClCI}{}\renewcommand{\ParaSampClCI}{[+3.2, +18.0]}
\providecommand{\ParaSampSingleClCI}{}\renewcommand{\ParaSampSingleClCI}{[-1.6, +14.0]}
\providecommand{\OftGoalMedoidClCI}{}\renewcommand{\OftGoalMedoidClCI}{[-8.4, -0.8]}
\providecommand{\OftGoalParaMedoidClCI}{}\renewcommand{\OftGoalParaMedoidClCI}{[-1.2, +8.4]}

}{}
\IfFileExists{tables/para_features_nums.tex}{\providecommand{\BaseTrigramRho}{}\renewcommand{\BaseTrigramRho}{+0.36}

\providecommand{\BaseEmbedRho}{}\renewcommand{\BaseEmbedRho}{+0.47}

\providecommand{\BaseHumanRho}{}\renewcommand{\BaseHumanRho}{0.00}

\providecommand{\BaseParaN}{}\renewcommand{\BaseParaN}{210}

}{}
\providecommand{\OftObjectFeasWrong}{\pending{OftObjectFeasWrong}}
\IfFileExists{tables/probe_feasibility_nums.tex}{

\providecommand{\OftObjectFeasShare}{}\renewcommand{\OftObjectFeasShare}{1}
\providecommand{\OftObjectFeasWrong}{}\renewcommand{\OftObjectFeasWrong}{96.0}
\providecommand{\BaseObjectFeasShare}{}\renewcommand{\BaseObjectFeasShare}{98}
\providecommand{\BaseObjectFeasWrong}{}\renewcommand{\BaseObjectFeasWrong}{2.0}

}{}

\title{Paraphrase Consensus Is Not Self-Consistency:\\Instruction Reliance and Robustness in VLA Policies}
\author{Hanlong Li \quad Sihan Tan \quad Takeshi Ashizawa \quad Benjamin Yen \quad Kazuhiro Nakadai \\
  Institute of Science Tokyo, Tokyo, Japan}
\hypersetup{pdftitle={Paraphrase Consensus Is Not Self-Consistency: Instruction Reliance and Robustness in VLA Policies},
  pdfauthor={Hanlong Li, Sihan Tan, Takeshi Ashizawa, Benjamin Yen, Kazuhiro Nakadai}}
\date{}

\begin{document}
\maketitle

\begin{abstract}
Vision-language-action (VLA) policies fine-tuned on LIBERO are trained and evaluated on the same ten instruction strings, and rewording the instruction changes base OpenVLA's success by $\UntouchedCost$ points. Test-time prompt ensembling, which improves language models, does not recover this loss in any setting we ran. Paraphrase consensus (the medoid action over $K$ paraphrases) costs base OpenVLA $\BaseMedoidMag$ points when the training instruction is a member. Without it, the medoid of the paraphrases scores $\ParaMedoidVsCanonDelta$ points relative to the training instruction, and confidence-based selection scores $\ParaConfVsCanonDelta$. The same medoid over samples of one paraphrase improves on greedy decoding but stays $\ParaSampMedoidVsCanonDelta$ points below the training instruction, consistent with aggregation reducing variance without removing the bias the paraphrases share. A wrong-instruction probe measures how much of a policy's success depends on the instruction. Across three policies and four suites, paraphrases cost a policy only where this share is high, and the stronger policies lose a much smaller fraction of it than base OpenVLA. LoRA post-training on paraphrased instructions reduces the one-paraphrase effect to $\AugCost$ points, against $\CtrlCost$ for a matched control, reproduces across two seeds and on OpenVLA-OFT, transfers to withheld phrasings, and leaves the policy's reliance on the instruction unchanged.
\end{abstract}

\section{Introduction}
\label{sec:intro}
Vision-language-action (VLA) policies map a camera image and a natural-language instruction to robot actions~\citep{brohan2023rt1,zitkovich2023rt2,kim2024openvla,black2024pi0}. Many are language models fine-tuned to emit actions. OpenVLA~\citep{kim2024openvla}, for example, is a 7B Llama-2 model that outputs actions as vocabulary tokens. On LIBERO~\citep{liu2023libero}, the standard simulation benchmark for these policies, each released checkpoint is fine-tuned on exactly ten instructions, one fixed string per task, and evaluated on the same ten strings, character for character~\citep{kim2025oft}, which in NLP terms is an extreme case of single-template instruction tuning. Users will not phrase a request exactly as the training data did, and language models are sensitive to how a prompt is worded~\citep{sclar2024quantifying,lu2022fantastically,sun2024evaluating,mizrahi2024state}, so a benchmark score says little about what happens when the instruction is reworded. We ask how much rewording costs such a policy, and whether that loss can be recovered at test time, without retraining, the way prompt ensembling and self-consistency recover accuracy for language models.

The natural test-time remedy comes from language models. Self-consistency~\citep{wang2023selfconsistency} samples several answers and keeps the majority, and prompt ensembling~\citep{jiang2020know} aggregates answers over rewordings of a query. The VLA analogue is \emph{paraphrase consensus}: rephrase the instruction $K$ ways, query the policy once per phrasing, and execute the medoid, the action closest to all the others, a continuous majority vote that always returns one of the policy's own actions. It needs no training or verifier, and test-time selection over sampled actions already helps VLAs~\citep{kwok2025robomonkey,jang2026mgselect}. Prompt ensembling, however, works for language models pre-trained on many phrasings, where rewordings are comparably good prompts. A VLA fine-tuned on one string per task has seen only one good prompt, and every paraphrase may be a worse one. We implement paraphrase consensus (\method{}, \methodlong) with paired, per-step logging and run it on OpenVLA-OFT (LIBERO-Spatial with and without a distractor object, and Goal) and base OpenVLA (Spatial and Long). For these two policies and $\pi_{0.5}$ on all four suites, we also measure how much of each policy's success needs the instruction at all. We find that unweighted consensus does not help. The per-step logs trace the failure to the members themselves, which points to a fix at training time.

\textbf{Test-time consensus does not recover what a paraphrase costs.} One paraphrase, held for the episode, changes base OpenVLA's success by $\UntouchedCost$ points. Consensus with the training instruction among the members costs base OpenVLA $\BaseMedoidMag$ points and OFT $\OftGoalMedoidMag$ on Goal; without it, neither the medoid of the paraphrases nor the paraphrase the policy is most confident in comes close to the training instruction ($\ParaMedoidVsCanonDelta$, $\ParaConfVsCanonDelta$). The reason is that paraphrases are not exchangeable. Self-consistency aggregates $K$ draws from \emph{one} conditional, whereas $K$ paraphrases are $K$ conditionals, only one of which the policy was fit to. The same medoid over samples of one paraphrase does reduce variance, beating greedy decoding by $\ParaSampMedoidVsGreedyDelta$, yet stays $\ParaSampMedoidVsCanonDelta$ below the training instruction (\S\ref{sec:weak}).

\textbf{A paraphrase costs a policy only where its success needs the instruction.} A wrong-instruction probe, a partial-input baseline~\citep{poliak2018hypothesis,gururangan2018annotation,kaushik2018much,thomason2019shifting}, gives another task's instruction in the same scene. Base OpenVLA needs the instruction for nearly all of its success on every suite. OpenVLA-OFT and $\pi_{0.5}$~\citep{intelligence2025pi05} pay a paraphrase cost only on LIBERO-Goal, where all of their success needs the instruction, and lose a far smaller fraction of it than base OpenVLA does (\S\ref{sec:probe}). This qualifies LIBERO-Plus's conclusion that fine-tuned VLAs ``degenerate into vision-action models''~\citep{fei2025liberoplus}.

\textbf{The fix is at training time.} LoRA post-training on the same demonstrations with paraphrased instructions moves the one-paraphrase effect from $\UntouchedCost$ to $\AugCost$, while a matched control trained on the original instructions stays at $\CtrlCost$. The result holds for a second seed, transfers to phrasings withheld from training, carries to OpenVLA-OFT on Goal, and leaves the policy reading its instruction. It costs about an hour of training once, whereas consensus queries four instructions at every step (\S\ref{sec:aug}).

\textbf{Contributions.} (i)~A paired, per-timestep measurement of paraphrase consensus on two policies in five settings, with the mechanism read from the logs and tested by an exchangeable-sampling control and a confidence-weighted selector. (ii)~An instruction-reliance probe on all LIBERO suites, a partial-input baseline for embodied models, and the finding that paraphrase cost appears only where reliance is high, across twelve policy--suite pairs from three policies. (iii)~The established remedy, training on paraphrases~\citep{xiao2022dial}, measured with three controls we did not find reported for prior remedies~\citep{hou2026langgap,yin2026gsr}: a matched training control, transfer to withheld phrasings and types, and a grounding check.

\section{Related Work}
\label{sec:related}
\textbf{VLAs and their language robustness.} Large pretrained policies map images and language to actions~\citep{brohan2023rt1,zitkovich2023rt2,ghosh2024octo,kim2024openvla,black2024pi0,intelligence2025pi05}. OpenVLA-OFT adds parallel decoding, a continuous action head and chunking~\citep{kim2025oft}. LIBERO's perturbation suites reach opposite conclusions on language: LIBERO-Plus finds fine-tuned VLAs largely ignore the instruction~\citep{fei2025liberoplus}, LIBERO-PRO reports collapse under minimal rewording~\citep{zhou2025liberopro}, and LIBERO-Para quantifies the collapse at $22$--$52$ points under paraphrase across seven VLAs, attributed to surface-level matching rather than semantic grounding~\citep{kim2026liberopara}. \citet{xing2025shortcut} attribute language-ignoring to shortcut learning, and LangGap shows $\pi_{0.5}$ largely ignores instructions when one task per layout lets it~\citep{hou2026langgap}. LIBERO-Para is built on LIBERO-Goal, where its paraphrases cost OFT about $33$ points. Ours cost it $\OftGoalUntouchedCost$ there and base OpenVLA $\GoalUntouchedCost$. Our generator preserves every object reference, and LIBERO-Para attributes most of its drop to object-level lexical variation. What these diagnostics do not ask is whether test-time aggregation recovers the loss, or how the penalty relates to how much of a score needs the instruction at all (\S\ref{sec:probe}).

\textbf{Perturbation ensembles and failure detection.} Ensemble disagreement~\citep{lakshminarayanan2017ensembles,ovadia2019trust,menda2019ensembledagger,hoque2021thriftydagger,romer2025fiper,lee2026pfd,shanmugam2021tta} and runtime monitors~\citep{agia2024stac,gu2025safe,seligmann2026vlafail,karli2025askbefore} are compared with paraphrase on identical trajectories in App.~\ref{sec:sensitivity} and \ref{app:failure}.

\textbf{Prompt sensitivity, self-consistency and test-time scaling.} LLMs are sensitive to how a prompt is worded: formatting~\citep{sclar2024quantifying}, example order~\citep{lu2022fantastically} and paraphrased instructions~\citep{sun2024evaluating,mizrahi2024state} move accuracy by many points. Prompt ensembling~\citep{jiang2020know}, self-consistency~\citep{wang2023selfconsistency}, semantic entropy~\citep{kuhn2023semantic,farquhar2024semantic} and paraphrase-based uncertainty~\citep{gao2024spuq,khan2024consistency} aggregate over samples or prompts in discrete output spaces. The medoid is minimum-Bayes-risk selection, of which self-consistency is a special case~\citep{bertsch2023mbr}, and MBR over several prompts' outputs helps text generation~\citep{heineman2024mbr}. For VLAs, RoboMonkey scores sampled actions with a verifier~\citep{kwok2025robomonkey} and MG-Select scores against a condition-masked reference~\citep{jang2026mgselect}, and both sample the \emph{action} axis from one prompt. Paraphrase consensus samples the \emph{instruction} axis. Aggregators built for prompts of unequal quality weight members by held-out accuracy~\citep{jiang2020know,arora2023ask} or zero-shot scores~\citep{allingham2023zeroshot}, and lower-perplexity prompts work better~\citep{gonen2023demystifying}. With one member known to be the training instruction, that weighting degenerates to selecting it, which is the canonical controller. \S\ref{sec:weak} shows the difference between the two axes is decisive.

\textbf{Remedies.} Training on more phrasings is an established remedy. DIAL relabels demonstrations with VLM-generated instructions to broaden what a policy can follow~\citep{xiao2022dial}, SPRINT relabels for pre-training~\citep{zhang2023sprint}, and RT-1 trained on diverse phrasings~\citep{brohan2023rt1}. LangGap closes its language gap with targeted training-time augmentation~\citep{hou2026langgap}, and GSR argues the brittleness is architectural and retrains an action expert on separately extracted task semantics~\citep{yin2026gsr}. In NLP, multi-template instruction tuning~\citep{wei2022finetuned,sanh2022multitask}, consistency training under back-translation paraphrases~\citep{xie2020unsupervised}, and prompt- and paraphrase-consistency objectives~\citep{zhou2022prompt,elazar2021measuring,sun2024evaluating} are the precedents for training on rewordings, and our paraphrase types are CheckList-style invariance tests~\citep{ribeiro2020beyond} for an embodied model. \S\ref{sec:aug} adds three measurements we did not find in these works: a matched training control that separates the phrasings from the extra steps, transfer to withheld phrasings, and a check that the remedy does not work by ignoring the instruction.

\section{Paraphrase Consensus}
\label{sec:method}
\textbf{Setting.} At replanning step $t$ a frozen policy $\pi$ maps observation $o_t$ and instruction $\ell$ to an action chunk $A\in\mathbb{R}^{H\times D}$; receding-horizon control executes $h\le H$ steps and re-queries. We are given $\mathcal{L}=\{\ell_1,\dots,\ell_K\}$ with $\ell_1$ the training instruction (the benchmark's canonical string) and $\ell_{2..K}$ semantically equivalent paraphrases. Including $\ell_1$ makes this the most favourable test of test-time computation over phrasings, and \S\ref{sec:weak} also runs the deployable ensemble without it. $\pi$ is not modified, and only what is queried and what is executed changes.

\textbf{Paraphrase sets.}
\label{sec:paraphrase}
For each task an LLM generates candidates instructed to preserve every object, attribute, spatial relation and goal while varying lexical choice and syntax. Candidates are kept if an independent LLM judge rates them equivalent and their sentence-embedding cosine to the training instruction exceeds a threshold, and the final sets are chosen by greedy max--min selection on embedding distance. Per task we release an \emph{llm} set, \emph{easy} (lexical substitution), \emph{hard} (structural rewrite), \emph{adversarial} (typos and casing) and a \emph{held-out} set never used at training time (examples in Tab.~\ref{tab:example}). Prompts, judge, embedding model and threshold are in App.~\ref{app:paraphrases}. $K{=}4$ unless stated. One author, blind to type and suite though not to the hypotheses, judged \HumanDiff{} of the \HumanN{} evaluated paraphrases (\HumanDiffPct\%) to change the task although the LLM judge passed them. On the pairs judged equivalent every one-paraphrase cost below stays inside its original $95\%$ interval (App.~\ref{app:human}).

\textbf{Disagreement.} Dimensions are scaled by the checkpoint's own per-dimension action statistics $s\in\mathbb{R}^D$ (those used to un-normalise its outputs), and
\begin{align}
 d(A,B)&=\tfrac{1}{H}\sum_{h=1}^{H}\bigl\lVert (A_{h,:}-B_{h,:})\oslash s\bigr\rVert_2, \label{eq:u}\\
 \ut&=\tfrac{2}{K(K-1)}\sum_{i<j} d\bigl(A^{(i)}_t,A^{(j)}_t\bigr),\nonumber
\end{align}
with $A^{(k)}_t=\pi(o_t,\ell_k)$.

\textbf{Consensus action.}
\label{sec:medoid}
We execute the medoid, $k^\star=\argmin_k\sum_{j\ne k} d(A^{(k)}_t,A^{(j)}_t)$, ties toward $\ell_1$. It is always a chunk the policy produced (the mean of two grasps from opposite sides is a collision), a single divergent member cannot pull it, and under a $0/1$ metric on whole actions it would be the mode, i.e.\ self-consistency voting~\citep{wang2023selfconsistency}. With the $L_2$ metric of Eq.~\eqref{eq:u} it is a continuous relaxation of that vote. Controls hold the aggregation fixed and vary the members: the mean, a random single paraphrase, the medoid of the paraphrases alone, and (\S\ref{sec:weak}) $K$ stochastic decodes of $\ell_1$.

\textbf{Why aggregating over prompts is not variance reduction.} Write the $K$ members as $A^{(k)}=a^\star+b_k+\varepsilon_k$, where $a^\star$ is a task-correct action, $b_k$ the systematic bias of conditioning on $\ell_k$, and $\mathbb{E}\,\varepsilon_k=0$. For the mean aggregator,
\begin{align}
\mathbb{E}\bigl\lVert \bar A - a^\star\bigr\rVert^2 &= \Bigl\lVert \tfrac{1}{K}\textstyle\sum_k b_k\Bigr\rVert^2 \label{eq:biasvar}\\
&\quad+ \tfrac{1}{K^2}\textstyle\sum_k \mathbb{E}\lVert\varepsilon_k\rVert^2 .\nonumber
\end{align}
Self-consistency is the exchangeable case $b_k\equiv b_1$: the bias term is that of the single prompt whatever $K$ is, and the variance term falls as $1/K$, so aggregation can only help. Paraphrase consensus replaces $b_1$ by the \emph{average} bias of $K$ different conditionings. That average beats a single query with $\ell_1$ only if the paraphrase biases are no larger than $b_1$ or cancel; the decomposition does not say whether they do, and \S\ref{sec:weak} measures it: the cost of executing one paraphrase shows how much larger they are, and the medoid shows they do not cancel, while samples of one paraphrase show the variance term is real. The medoid is not the mean, but it selects among the same members, so the question is about the members.

\textbf{Cost.} The shared visual prefix does not make the $K$ prompts cheap: measured for OFT on an idle A100, $K{=}4$ costs $\LatencyKFourRatio\times$ one query even batched (App.~\ref{app:cost}), so query count is a fair unit of cost.

\section{Experiments}
\label{sec:exp}

\subsection{Setup}
\label{sec:setup}
\textbf{Policies.} OpenVLA-OFT with the public per-suite LIBERO checkpoints~\citep{kim2025oft} ($H{=}8$, L1 regression head, deterministic), and base OpenVLA with its per-suite LIBERO checkpoints~\citep{kim2024openvla} (one action per query, 256 discrete bins per dimension, so $H{=}1$ and the policy can be sampled). Both are 7B models on the same backbone and data, and OFT additionally consumes a wrist camera and proprioception. For the reliance and paraphrase-cost measurements only (\S\ref{sec:probe}) we add $\pi_{0.5}$~\citep{intelligence2025pi05}, its released LIBERO checkpoint run through openpi's own client (App.~\ref{app:repro}).

\textbf{Benchmarks.} The four LIBERO suites (10 tasks each, fixed initial states) and a distribution-shifted LIBERO-Spatial from LIBERO-PRO~\citep{zhou2025liberopro} that adds a distractor mug, on which OFT falls from $\OftStdCanonRate\%$ to $\OftShiftCanonRate\%$. Both policies reproduce their published numbers through our harness (OFT within a $\pm2$-point gate; base OpenVLA within its Wilson intervals; App.~\ref{app:repro}). Every comparison is paired: the same initial states in the same order, $50$, $25$ or $10$ episodes per task, as stated per table.

\textbf{Controllers.} \emph{Canonical-$h$}: one query with the training instruction, fixed horizon $h$ (Canonical-8 for OFT, Canonical (greedy) for base OpenVLA, where $h{=}1$). \emph{One paraphrase}: one of the $K{-}1$ paraphrases, never the training instruction, fixed for the episode. \emph{One of $K$ instructions}: the same draw with the training instruction included, per episode or per replan. \emph{Mean ensemble}: mean of the $K$ chunks. \emph{Paraphrase medoid} (\method): the medoid of the $K$ chunks. \emph{Paraphrase-only medoid}: the medoid of the $K{-}1$ paraphrases' chunks, $\ell_1$ never queried. \emph{Sampling single / medoid}: one or $K$ stochastic decodes of the training instruction (base OpenVLA only; \S\ref{sec:weak}). Adaptive-horizon, gated, monitor-only and matched-random-horizon variants were also run on standard LIBERO, and all are null there (App.~\ref{app:std}--\ref{app:horizon}).

\textbf{Statistics.} Success with Wilson $95\%$ intervals~\citep{wilson1927}; paired $\Delta$ against the canonical reference with $95\%$ bootstrap intervals ($10{,}000$ resamples over episodes) and exact McNemar tests~\citep{mcnemar1947}, Holm-corrected within each table. Comparisons quoted only in the text are single tests. Two runs of one OFT configuration on the same $500$ initial states disagree on $\HarnessDiscPct\%$ of episodes (bfloat16 kernels under load), so the minimum detectable paired effect at $n{=}500$ is $\MdePts$ points, comparable to the headroom Canonical-8 leaves on standard LIBERO-Spatial. Unintended replicates of the baseline (a routing bug, App.~\ref{app:repro}) differ from it by $\OftReplDiff$ points on OFT and $\BaseReplDiff$ on base OpenVLA: the noise floor for a paired $\Delta$, which the episode-level intervals do not include. Intervals that also resample tasks (App.~\ref{app:clustered}) are wider and leave the sign of every headline effect unchanged.

\subsection{Consensus on a strong policy}
\label{sec:main}
On Spatial, with and without shift, consensus is null for the stronger policy: every controller compared is within $\OftStdMaxAbsDelta$ points of Canonical-8 on standard LIBERO-Spatial, and on the shifted suite, which has $\OftShiftHeadroom$ points of headroom, the medoid reaches $\OftShiftMedoidRate\%$ against $\OftShiftCanonRate\%$ ($\OftShiftMedoidDelta$ $\OftShiftMedoidCI$, $p_{\text{Holm}}{=}\OftShiftMedoidPHolm$, $n{=}\OftShiftMedoidN$; App.~\ref{app:std}, Tab.~\ref{tab:main-shift}). On LIBERO-Goal, where one paraphrase costs OFT $\OftGoalUntouchedCost$ (\S\ref{sec:probe}), consensus with the training instruction costs $\OftGoalMedoidDelta$ $\OftGoalMedoidCI$ against querying it alone (task-clustered $\OftGoalMedoidClCI$), and the paraphrase-only medoid gains $\OftGoalParaMedoidVsOneDelta$ $\OftGoalParaMedoidVsOneCI$ over one paraphrase ($p{=}\OftGoalParaMedoidVsOneP$; task-clustered $\OftGoalParaMedoidClCI$), staying $\OftGoalParaMedoidDelta$ below it.

\subsection{Consensus on a weak policy: a large harm, and its mechanism}
\label{sec:weak}
Base OpenVLA on LIBERO-Spatial has $\BaseHeadroom$ points of headroom and no distribution shift, the regime in which consensus has room to help, and consensus costs it $\BaseMedoidMag$ points ($\BaseMedoidRate\%$, $\BaseMedoidPairedCI$, $p<0.001$, with $\BaseMedoidDiscPct\%$ of paired episodes changing outcome). On LIBERO-Long the change is $\BaseLongMedoidDelta$ $\BaseLongMedoidCI$ ($n{=}\BaseLongMedoidN$): the same sign, not significant. Tab.~\ref{tab:weak-main} places the controls beside it.

\begin{widetable}
\centering\small
\caption{Base OpenVLA on LIBERO-Spatial, $n{=}250$ per arm, paired. The replicate row is a second run of the baseline under a mis-routed controller name and measures run-to-run noise. Sampling rows use stochastic decodes of the training instruction with the same medoid code. Generated by \texttt{lcc/analysis/main\_table.py}.}
\label{tab:weak-main}
\begin{tabular}{lrrlr}
\toprule
Method & $n$ & Succ.\ (\%) & $\Delta$ vs.\ ref.\ [95\% CI] & $p_{\text{Holm}}$ \\
\midrule
\multicolumn{5}{l}{\textit{base OpenVLA / LIBERO-Spatial, reference: Canonical (greedy)}} \\
\midrule
Canonical (greedy) & 250 & 85.6 \tiny[80.7, 89.4] & --- & --- \\
Canonical (replicate run) & 250 & 83.6 \tiny[78.5, 87.7] & -2.0 [-5.6, 1.2] & 0.888 \\
Paraphrase medoid, $K{=}4$ & 250 & 68.8 \tiny[62.8, 74.2] & \textbf{-16.8 [-22.8, -10.8]} & $<$0.001 \\
One paraphrase, fixed per episode & 250 & 47.6 \tiny[41.5, 53.8] & \textbf{-38.0 [-44.8, -31.2]} & $<$0.001 \\
Paraphrase-only medoid, $K{-}1{=}3$ (no training instr.) & 250 & 53.2 \tiny[47.0, 59.3] & \textbf{-32.4 [-39.2, -25.6]} & $<$0.001 \\
One of $K{=}4$ instructions (canonical incl.), per episode & 250 & 58.4 \tiny[52.2, 64.3] & \textbf{-27.2 [-34.0, -20.4]} & $<$0.001 \\
One of $K{=}4$ instructions (canonical incl.), per replan & 250 & 66.4 \tiny[60.3, 72.0] & \textbf{-19.2 [-25.6, -13.2]} & $<$0.001 \\
Sampling medoid, $K{=}4$ [T=0.5] & 250 & 88.4 \tiny[83.8, 91.8] & +2.8 [-1.6, 7.2] & 0.888 \\
Sampling medoid, $K{=}4$ [T=1] & 250 & 86.4 \tiny[81.6, 90.1] & +0.8 [-4.4, 6.0] & 0.888 \\
One stochastic decode [T=1] & 250 & 82.0 \tiny[76.8, 86.3] & -3.6 [-8.8, 1.6] & 0.888 \\
\bottomrule
\end{tabular}

\end{widetable}

\textbf{The mechanism is in the logs.} The medoid is the member minimising total distance to the others, so it returns the training instruction's action only when that action is central. For base OpenVLA it is not. Its paraphrase disagreement is $u{=}\UntouchedU$ per replan against OFT's $\OFTStdU$ on the same suite and scale (indicative only, since OFT's $u$ averages an $8$-step chunk), and the selected index is nearly uniform over the four members ($\UntouchedHist$). On $\UntouchedParaFrac\%$ of replans the executed action is conditioned on a paraphrase, and on a different one from step to step.

\textbf{Decomposition.} Tab.~\ref{tab:weak-main} separates the members from the aggregation. Conditioning on one paraphrase, held for the episode, changes success by $\UntouchedCost$ points (\UntouchedPara\% against \UntouchedCanon\%). Letting the per-episode draw include the training instruction (one of the $K{=}4$ instructions, so a quarter of the episodes pay nothing) costs $\OneOfKEpisodeMag$; redrawing at every replan costs $\OneOfKReplanMag$ (a bad draw then costs one step, not the episode); the medoid of all four costs $\BaseMedoidMag$. Aggregation returns part of what a paraphrase costs and far from all of it, and a selector could recover the rest only by identifying the training instruction's member, which amounts to the canonical controller (a confidence-based selector is tested below). The penalty is sharply task-dependent (App.~\ref{app:pertask}). $K{=}8$ gives $\BaseKEightDelta$ $\BaseKEightCI$ ($n{=}\BaseKEightN$), an interval that includes the $K{=}4$ cost; $K{=}2$ is the canonical controller by the tie rule (App.~\ref{app:horizon}).

\textbf{The deployable ensemble.} A deployed system has the user's phrasing, not $\ell_1$. Querying the $K{-}1{=}3$ paraphrases alone, with the training instruction never queried, the medoid scores \ParaMedoidRate\%, not distinguishable from a single paraphrase ($\ParaMedoidVsOneDelta$ $\ParaMedoidVsOneCI$, $p{=}\ParaMedoidVsOneP$; task-clustered $\ParaMedoidClCI$, so a moderate gain is not excluded), and $\ParaMedoidVsCanonDelta$ $\ParaMedoidVsCanonCI$ below the training instruction (Tab.~\ref{tab:weak-main}). Greedy members carry no sampling noise, so what the medoid gains here is paraphrase-specific error cancelling, while the bias the paraphrases share remains; the earlier medoid owed its performance to the training-instruction member, and the paraphrase-only medoid scores $\ParaMedoidVsCanonMedoidDelta$ against it. Weighting by the policy's own confidence, as perplexity-based prompt selection suggests~\citep{gonen2023demystifying}, does not close the gap: executing at each replan the paraphrase whose greedy action has the highest mean action-token log-probability scores \ParaConfVsOneRate\%, $\ParaConfVsOneDelta$ $\ParaConfVsOneCI$ over one paraphrase (task-clustered $\ParaConfVsOneClCI$) and $\ParaConfVsCanonDelta$ $\ParaConfVsCanonCI$ below the training instruction. Given the training instruction as a fourth member, confidence picks it on \ConfCanonFrac\% of replans (uniform $25\%$, medoid \ConfMedoidCanonFrac\%) and costs $\ConfVsCanonDelta$ $\ConfVsCanonCI$ ($n{=}\ConfVsCanonN$): the policy's confidence barely recognises the instruction it was trained on.

\textbf{The exchangeable control.} The argument predicts what happens if the members are made exchangeable: replace the $K$ paraphrase responses by $K{=}4$ stochastic decodes of the \emph{training} instruction (temperature $1.0$, action vocabulary only) and aggregate with the identical medoid. Neither a single stochastic decode ($\SamplingSingleDelta$ $\SamplingSingleCI$ vs.\ greedy) nor the medoid of four ($\SamplingMedoidDelta$ $\SamplingMedoidCI$) is distinguishable from greedy (Tab.~\ref{tab:weak-main}, Fig.~\ref{fig:2x2}). Paired on the same episodes, the sampling medoid is $\SamplingVsParaMedoidDelta$ $\SamplingVsParaMedoidCI$ above the paraphrase medoid. The harm therefore comes from the members rather than from the aggregation step. Around $\ell_1$ there is little variance to remove. Around a paraphrase there is more. Four stochastic decodes of the episode's paraphrase, aggregated by the identical medoid, score \ParaSampMedoidRate\%: $\ParaSampMedoidVsGreedyDelta$ $\ParaSampMedoidVsGreedyCI$ above greedy decoding of that paraphrase (task-clustered $\ParaSampClCI$) and $\ParaSampMedoidVsSingleDelta$ $\ParaSampMedoidVsSingleCI$ above a single stochastic decode of it (task-clustered $\ParaSampSingleClCI$), but $\ParaSampMedoidVsCanonDelta$ $\ParaSampMedoidVsCanonCI$ below the training instruction. Exchangeable members reduce variance, which is self-consistency's guarantee. No aggregation we ran removes the bias the paraphrases share, the first term of Eq.~\eqref{eq:biasvar}.

\subsection{Instruction reliance: how much of a score needs the instruction?}
\label{sec:probe}
A policy robust to rewording and a policy not reading the instruction look alike under paraphrase. LIBERO-Plus, which blanks the instruction, concludes fine-tuned VLAs ``degenerate into vision-action models''~\citep{fei2025liberoplus}. A blank prompt cannot separate the two readings.

We instead condition the policy on a \emph{different task's} training instruction (task $i{+}5$ of the same suite) while the scene, the initial state, and the success predicate remain task $i$'s. Success then means the policy did not need the instruction. Unlike a blanked prompt, the wrong instruction is fluent and in-distribution, so ``ignores language'' is not confounded with ``falls back on a prior when the text is garbage''. Unlike LIBERO-Plus's goal-replacement experiment, which changes the success predicate too and therefore measures cross-object generalisation, this measures reliance. Every cell is a single canonical query, paired with its own correct-instruction reference on the same initial states. Tab.~\ref{tab:reliance} (App.~\ref{app:post}; per-cell rates in Tab.~\ref{tab:probe}) reports the language share, $(\text{correct}-\text{wrong})/\text{correct}$: the fraction of the score that requires reading the instruction (an upper bound, since a fluent wrong instruction can also misdirect).

\paragraph{Reliance depends on the suite.}
OpenVLA-OFT needs the instruction for $\OftGoalShare\%$ of its success on LIBERO-Goal and for $\OftObjectShare\%$ on LIBERO-Object, and the task definitions say why. Spatial's ten tasks differ only in a referring expression over one object set with two black bowls, so a policy that ignores the instruction still picks the right bowl $\OftSpatialWrongPct\%$ of the time. Goal's ten tasks are different objectives in one scene, and nothing but the instruction identifies which. Told the wrong one, all three policies score exactly $0$. Object's tasks name different targets, yet OFT retrieves the right one $\OftObjectWrongPct\%$ of the time when told the wrong name: the scene identifies the task, as LIBERO-Plus saw with a blank prompt. This holds where the wrong instruction is executable, and on Long, where none is, the share measures success when the instruction cannot be followed (App.~\ref{app:probefeas}). A LIBERO aggregate thus mixes suites where language carries $\OftObjectShare\%$ of the task with suites where it carries all of it, and ``VLAs ignore language'' is true of some suites and false of others for the same checkpoint.

\paragraph{Reliance also depends on the policy, and paraphrase cost appears only where it is high.}
Base OpenVLA needs the instruction for $\BaseSpatialShare$--$\BaseGoalShare\%$ of its success on every suite, OpenVLA-OFT only where the benchmark forces it. A paraphrase that conditions the policy no worse than another task's instruction can cost at most what the instruction carries, $\text{share}\times\text{correct}$, so a low share bounds the cost by construction. The informative questions are how much of the bound a paraphrase uses and whether a high share implies a cost. Across the twelve policy--suite pairs (\IfFileExists{tables/reliance_vs_cost.tex}{Tab.~\ref{tab:reliance}}{App.~\ref{app:post}}) it is: OFT's only cost distinguishable from zero is on Goal ($\OftGoalParaDelta$ $\OftGoalParaCI$, share \OftGoalShare\%), and base OpenVLA, at $\BaseSpatialShare\%$ share or more everywhere, changes by $\BaseObjectParaDelta$ to $\BaseSpatialParaDelta$. $\pi_{0.5}$, a PaliGemma-based flow-matching policy from a different family, has shares of \PiSpatialShare, \PiObjectShare, \PiGoalShare{} and \PiLongShare\% and again pays a cost distinguishable from zero only on Goal ($\PiGoalParaDelta$ $\PiGoalParaCI$), and both statements hold after Holm correction across the twelve cells. Per-task correlations are in App.~\ref{app:reltask}. As a fraction of the success the instruction carries, a paraphrase removes \OftGoalLost\% (OFT) and \PiGoalLost\% ($\pi_{0.5}$) on Goal against \BaseSpatialLost\% and \BaseGoalLost\% for base OpenVLA on Spatial and Goal, with non-overlapping intervals (Tab.~\ref{tab:reliance}): the stronger policies read the instruction there and still follow most rewordings. A high share is not sufficient for a cost, nor does it set its size: $\pi_{0.5}$ relies on the instruction for \PiObjectShare\% of its Object success while a paraphrase changes it by $\PiObjectParaDelta$, and at $\PiGoalShare\%$ share on Goal the change is $\BaseGoalParaDelta$ for base OpenVLA and $\OftGoalParaDelta$ for OFT. A failing partial-input baseline does not by itself prove the full model uses the input~\citep{feng2019misleading}.

\section{The Training-Time Remedy: Instruction Augmentation}
\label{sec:aug}
\begin{widetable}
\centering\small
\caption{Instruction augmentation on base OpenVLA / LIBERO-Spatial (LoRA rank 32, learning rate $10^{-4}$, $2{,}000$ steps). Paraphrased instructions with ground-truth actions against a matched control trained on training instructions and the untouched checkpoint. All cells $n{=}250$, paired on identical initial states; held-out paraphrases were withheld from training. Generated by \texttt{lcc/distill/report\_p5.py}.}
\label{tab:aug}
\resizebox{0.84\floatwidth}{!}{\begin{tabular}{lrrrrl}
\toprule
Arm & Canonical & Paraphrase & Held-out par. & Wrong instr. & Paraphrase cost [95\% CI] \\
\midrule
untouched checkpoint & 85.6 & 47.6 & 57.6 & 2.4 & \textbf{$-38.0$ $[-44.8, -31.2]$} \\
augmentation, p=1 & 81.6 & 83.6 & 83.6 & 2.8 & $+2.0$ $[-4.0, +8.0]$ \\
canonical-only control & 82.8 & 50.8 & 56.4 & -- & \textbf{$-32.0$ $[-38.8, -25.2]$} \\
augmentation, p=0.5 & 82.4 & 78.4 & -- & -- & $-4.0$ $[-10.0, +2.0]$ \\
augmentation, lr 5e-5 & 83.6 & 82.8 & -- & -- & $-0.8$ $[-6.8, +5.2]$ \\
\bottomrule
\end{tabular}
}
\end{widetable}
If the training instruction is privileged because training only ever used it, the remedy is to widen the training distribution, not to average over it at test time. We post-train base OpenVLA with LoRA (rank 32, learning rate $10^{-4}$, $2{,}000$ steps) on its own LIBERO-Spatial demonstrations, keeping the ground-truth actions and replacing each sample's instruction with a paraphrase of the same task drawn from the \emph{llm}, \emph{easy}, and \emph{hard} splits (210 phrasings over 10 tasks). The \emph{held-out} split is withheld from training. A matched control is trained with the identical recipe, steps, and data on training instructions only: without it, a reduced paraphrase penalty is indistinguishable from the effect of more training. All four evaluation conditions are paired on the same 250 initial states (Tab.~\ref{tab:aug}). Other recipes are in App.~\ref{app:post}.

\paragraph{Augmentation removes the penalty.}
Executing a paraphrase changes the untouched checkpoint's success by $\UntouchedCost$ points and the control's by $\CtrlCost$ ($\CtrlCostCI$); after augmentation the change is $\AugCost$ ($\AugCostCI$), and $\AugSeedBCost$ against $\CtrlSeedBCost$ for a second training seed of both arms (Tab.~\ref{tab:augseed}). Because the control keeps the penalty, the change can be attributed to the instruction distribution rather than to the additional gradient steps. These paraphrases (the \emph{llm} set) were in the training pool. Transfer is next.

\paragraph{Transfer to withheld phrasings.}
On phrasings no arm saw, written in a different register (polite frames, filler openers), the untouched checkpoint pays $\UntouchedHeldoutCost$ $\UntouchedHeldoutCostCI$ and the control $\CtrlHeldoutCost$ $\CtrlHeldoutCostCI$; the augmented model scores \AugHeldout\%, $\AugHeldoutCost$ $\AugHeldoutCostCI$ relative to its own canonical rate (seed 2: $\AugSeedBHeldoutCost$, control $\CtrlSeedBHeldoutCost$). Paired arm against arm, augmentation beats the control by $\AugVsCtrlHeldDelta$ $\AugVsCtrlHeldCI$ on held-out phrasings and $\AugVsCtrlParaDelta$ $\AugVsCtrlParaCI$ on trained ones, at equal canonical success ($\AugVsCtrlCanonDelta$ $\AugVsCtrlCanonCI$). The held-out register is milder than the trained one ($\UntouchedHeldVsParaDelta$ $\UntouchedHeldVsParaCI$ untouched), so we assess transfer by these paired costs rather than by raw success rates.

\paragraph{Which rewordings hurt.} Breaking the paraphrase set by type (\IfFileExists{tables/variants.tex}{Tab.~\ref{tab:variants}}{App.~\ref{app:post}}; one paraphrase per episode, $n{=}250$ each) separates lexical substitution ($\UntouchedEasyDelta$ on the untouched policy), structural rewrites ($\UntouchedHardDelta$) and typos and casing ($\UntouchedAdversarialDelta$). Per paraphrase, base OpenVLA's cost shrinks with surface and embedding similarity to the training instruction ($\rho{=}\BaseTrigramRho$, $\BaseEmbedRho$; $n{=}\BaseParaN$), OFT's weakly and $\pi_{0.5}$'s not at all, and a human's equivalence label predicts none ($\rho{=}\BaseHumanRho$): the weaker policy's cost tracks the wording, though types partly drive the pooled correlation (App.~\ref{app:parafeat}). After augmentation, which trained on LLM, lexical and structural paraphrases, the changes are $\AugEasyDelta$, $\AugHardDelta$ and $\AugAdversarialDelta$. The largest residual is on the adversarial type, whose typos and casing were never trained (some of its items also reword, Tab.~\ref{tab:example}): it changes success by $\AugAdversarialDelta$ $\AugAdversarialCI$, not significant after Holm correction across the five types ($p_{\text{Holm}}{=}\AugAdversarialPHolm$) but the only type whose interval excludes zero.

\paragraph{The policy still reads the instruction.}
Invariance could come from ceasing to read, the failure LIBERO-Plus reports~\citep{fei2025liberoplus}. Under the wrong-instruction probe the augmented model's language share stays at \AugShare\% (wrong-instruction success $\UntouchedWrong\%$ to $\AugWrong\%$).

\paragraph{LIBERO-Goal, where only the instruction identifies the task.} On LIBERO-Goal, where every policy scores exactly $0$ under a wrong instruction (\S\ref{sec:probe}), the same recipe takes the paraphrase penalty from $\GoalUntouchedCost$ to $\GoalAugCost$ ($\GoalAugCostCI$), while the control, which never saw a paraphrase, retains $\GoalCtrlCost$ ($\GoalCtrlCostCI$). Canonical success does not fall ($\GoalAugCanonDelta$ $\GoalAugCanonCI$) and wrong-instruction success stays at $\GoalAugWrong\%$: the augmented policy still reads the instruction where nothing else identifies the task (Tab.~\ref{tab:auggoal}).

\paragraph{OpenVLA-OFT.} OpenVLA-OFT pays a paraphrase cost only on Goal (\S\ref{sec:probe}). The same recipe, run through OFT's own trainer (App.~\ref{app:post}), moves its one-paraphrase effect there from $\OftGoalUntouchedCost$ to $\OftGoalAugCost$ $\OftGoalAugCostCI$ and its held-out effect from $\OftGoalUntouchedHeldoutCost$ to $\OftGoalAugHeldoutCost$, while the matched control stays at $\OftGoalCtrlCost$ and $\OftGoalCtrlHeldoutCost$; the language share stays at \OftGoalAugShare\% (Tab.~\ref{tab:augoftgoal}). Paired arm against arm, augmentation beats the control by $\OftGoalAugVsCtrlParaDelta$ $\OftGoalAugVsCtrlParaCI$ on trained phrasings and $\OftGoalAugVsCtrlHeldDelta$ $\OftGoalAugVsCtrlHeldCI$ on held-out ones. Post-training itself lowers OFT's canonical success here, by $\OftGoalAugCanonDropDelta$ $\OftGoalAugCanonDropCI$ for augmentation and $\OftGoalCtrlCanonDropDelta$ $\OftGoalCtrlCanonDropCI$ for the control, so against the untouched model augmented paraphrase success is $\OftGoalAugCanonDropDelta$ rather than $\OftGoalAugCost$.

\paragraph{Cost to canonical success.}
On Spatial the augmented model's canonical-instruction success is $\AugCanon\%$ against the untouched $\UntouchedCanon\%$, and the control's $\CtrlCanon\%$: neither drop is distinguishable from zero ($\AugCanonDropDelta$ $\AugCanonDropCI$ and $\CtrlCanonDropDelta$ $\CtrlCanonDropCI$); the second seed gives $\SeedBAugCanonDropDelta$ and $\SeedBCtrlCanonDropDelta$ $\SeedBCtrlCanonDropCI$. Both arms drop by similar amounts, so the drop comes from post-training itself rather than from the paraphrases. The drop varies by recipe (Tab.~\ref{tab:augsweep}). An explicit symmetric-KL consistency term adds nothing measurable over augmentation (Tab.~\ref{tab:augcons}).

\paragraph{Consensus after augmentation.} If the members were the problem, the medoid should turn neutral once training makes them equally good conditioning. On the augmented checkpoint it scores $\AugMedoidDelta$ $\AugMedoidCI$ against its own canonical rate, disagreement falls from \UntouchedU{} to \AugU{} (indicative: different rollouts), and the medoid selects the training instruction's member \AugCanonFrac\% of the time (was \UntouchedCanonFrac\%). Once training has made the members agree, consensus is harmless but adds nothing measurable.

\paragraph{Mapping back to the training string.} Mapping the instruction back to the nearest training string presupposes the closed instruction set that deployment removes, and even on LIBERO it fails for a quarter of Spatial's structural rewrites (App.~\ref{app:canon}).

\ifacl\else
\section{Discussion and Limitations}
\label{sec:discussion}
\textbf{Limitations.} Consensus and post-training use two checkpoints of one 7B backbone, and the OFT/base contrast is confounded by OFT's wrist camera and proprioception. $\pi_{0.5}$ enters only the reliance and paraphrase-cost measurements, with $10$ episodes per task on Object and Long. One benchmark family, whose Spatial suite is $\OftSpatialWrongPct\%$ solvable without the instruction. Post-training has two seeds on base OpenVLA / Spatial and one on base OpenVLA / Goal and OpenVLA-OFT / Goal. It changes canonical success on Spatial by $\SeedBAugCanonDropDelta$ to $\SeedBCtrlCanonDropDelta$ points across arms and seeds (more for longer schedules), with only one control's bootstrap interval excluding zero, and significantly on OFT/Goal ($\OftGoalAugCanonDropDelta$), where the control pays about the same (\S\ref{sec:aug}, Tab.~\ref{tab:augseed}). Intervals are conditional on one run per arm: replicate runs of a baseline differ by $\OftReplDiff$--$\BaseReplDiff$ points, and two base-OpenVLA Long baseline runs by $\BaseLongReplDiff$ (App.~\ref{app:repro}). Every paraphrase, the held-out set included, is English, LLM-generated and LLM-judged by models from one vendor (App.~\ref{app:paraphrases}), so even the held-out register is LLM-written. One author's equivalence labels reject a third of the evaluated paraphrases without changing any reported cost (App.~\ref{app:human}), but they are one annotator's judgement, and human-written phrasings, which we have not collected, would be the stronger test of both penalty and remedy. The ensemble without the training instruction is built from paraphrases of that instruction, not of a user's own phrasing, and canonicalization was measured as retrieval accuracy, not in rollouts. Confidence-weighted selection was run on base OpenVLA / Spatial only. Each checkpoint saw ten instruction strings in fine-tuning, so the privilege of the training instruction may be weaker for VLAs trained on diverse language. No hardware results.
\fi

\section{Conclusion}
Unweighted paraphrase consensus over a frozen VLA did not improve success in any setting we ran and reduced it in two. Without the training instruction, both consensus and confidence-based selection stayed far below it ($\ParaMedoidVsCanonDelta$ and $\ParaConfVsCanonDelta$ points on base OpenVLA), because a paraphrase conditions the policy differently from the string it was trained on. The stronger policies are robust to rewording because most LIBERO suites do not require the instruction and, where one does, they follow most rewordings. Training on paraphrases removes the cost without making the policy ignore its instruction. Evaluating only on the training string is single-prompt evaluation~\citep{mizrahi2024state}, and we recommend reporting success on held-out paraphrases alongside it.\label{sec:endmain}

\ifacl
\section*{Limitations}
Consensus and post-training use two checkpoints of one 7B backbone, and the OFT/base contrast is confounded by OFT's wrist camera and proprioception. $\pi_{0.5}$ enters only the reliance and paraphrase-cost measurements, with $10$ episodes per task on Object and Long. One benchmark family, whose Spatial suite is $\OftSpatialWrongPct\%$ solvable without the instruction. Post-training has two seeds on base OpenVLA / Spatial and one on base OpenVLA / Goal and OpenVLA-OFT / Goal. It changes canonical success on Spatial by $\SeedBAugCanonDropDelta$ to $\SeedBCtrlCanonDropDelta$ points across arms and seeds (more for longer schedules), with only one control's bootstrap interval excluding zero, and significantly on OFT/Goal ($\OftGoalAugCanonDropDelta$), where the control pays about the same (\S\ref{sec:aug}, Tab.~\ref{tab:augseed}). Intervals are conditional on one run per arm: replicate runs of a baseline differ by $\OftReplDiff$--$\BaseReplDiff$ points, and two base-OpenVLA Long baseline runs by $\BaseLongReplDiff$ (App.~\ref{app:repro}). Every paraphrase, the held-out set included, is English, LLM-generated and LLM-judged by models from one vendor (App.~\ref{app:paraphrases}), so even the held-out register is LLM-written. One author's equivalence labels reject a third of the evaluated paraphrases without changing any reported cost (App.~\ref{app:human}), but they are one annotator's judgement, and human-written phrasings, which we have not collected, would be the stronger test of both penalty and remedy. The ensemble without the training instruction is built from paraphrases of that instruction, not of a user's own phrasing, and canonicalization was measured as retrieval accuracy, not in rollouts. Confidence-weighted selection was run on base OpenVLA / Spatial only. Each checkpoint saw ten instruction strings in fine-tuning, so the privilege of the training instruction may be weaker for VLAs trained on diverse language. No hardware results.
\fi

\ifacl
\section*{Ethics Statement}
This work uses public simulation benchmarks and public model checkpoints; no human participants or personal data are involved. LIBERO, LIBERO-PRO, OpenVLA and OpenVLA-OFT (code and checkpoints) are released under the MIT license, \texttt{transformers} and openpi under Apache~2.0, and the $\pi_{0.5}$ weights under the Gemma terms of use; all were used for research evaluation, their intended use, and our code is released under MIT. Paraphrases were generated and judged by an LLM (prompts in App.~\ref{app:paraphrases}), constrained to the benchmark's own objects and relations, and contain no personal data; no human phrasings were collected; the paraphrase equivalence labels of App.~\ref{app:human} were provided by an author. Two risks are practical. A physically acting policy can execute a different behaviour when a user rewords a command, and test-time consensus can make that worse; deployed systems should be evaluated under paraphrase, not only under the training string. And a robustness fix can mask instruction-following (a policy that stops reading its instruction); the wrong-instruction probe is designed to detect it and is reported for the principal trained models. Evaluation used \TotalEpisodes{} episodes, about \EvalProcessHours{} process-hours on shared A100s (roughly \EvalGPUHours{} GPU-hours); post-training took \TrainHours{} hours of wall-clock over \TrainRuns{} runs.
\else
\section*{Reproducibility Statement}
Every comparison is paired on the benchmark's fixed initial states, whose hashes are logged with each episode so that pairing is verified rather than assumed. Rollouts are logged at one row per replanning step (all $K$ predicted chunks, disagreement, selected member, latency), and every number in every table and figure is produced from those logs by the released analysis code, and numbers quoted in the prose come from those tables or from macro files the same code writes (\texttt{tables/*\_nums.tex}). Checkpoint revisions, environment versions, and the exact flags of the baseline reproduction are in App.~\ref{app:repro}, including a library-level hazard (a fork of \texttt{transformers} that removes causal masking) that silently changes base OpenVLA's behaviour and must be avoided to reproduce our baselines. Code, paraphrase sets with generation and judging prompts, per-timestep logs, and post-trained checkpoints are released.

\section*{AI Use Statement}
\fi

\bibliographystyle{\venuebibstyle}
\bibliography{references}

\appendix
\ifacl
\section{Reproducibility Statement}
\label{app:reproducibility}
Every comparison is paired on the benchmark's fixed initial states, whose hashes are logged with each episode so that pairing is verified rather than assumed. Rollouts are logged at one row per replanning step (all $K$ predicted chunks, disagreement, selected member, latency), and every number in every table and figure is produced from those logs by the released analysis code, and numbers quoted in the prose come from those tables or from macro files the same code writes (\texttt{tables/*\_nums.tex}). Checkpoint revisions, environment versions, and the exact flags of the baseline reproduction are in App.~\ref{app:repro}, including a library-level hazard (a fork of \texttt{transformers} that removes causal masking) that silently changes base OpenVLA's behaviour and must be avoided to reproduce our baselines. Code, paraphrase sets with generation and judging prompts, per-timestep logs, and post-trained checkpoints are released.
\fi
\section{Baseline Reproduction}
\label{app:repro}
The canonical controller is the \emph{unmodified} reference evaluation script, run at the published defaults (\texttt{use\_l1\_regression=True}, \texttt{use\_diffusion=False}, \texttt{use\_film=False}, \texttt{num\_images\_in\_input=2}, \texttt{use\_proprio=True}, \texttt{center\_crop=True}, \texttt{num\_open\_loop\_steps=8}, \texttt{num\_steps\_wait=10}, \texttt{lora\_rank=32}, \texttt{seed=7}), 50 episodes per task and 10 tasks per suite, on the fixed initial states the benchmark ships. Both tables in this section are generated from the evaluation logs by \texttt{lcc/analysis/repro.py}; no number in them is entered by hand.

\begin{widetable}
\centering\small
\caption{Reproduction of the published OpenVLA-OFT LIBERO results on the corrected stack. Intervals are Wilson 95\% intervals on our own rate~\citep{wilson1927}; the gate is $\pm 2$ points (500 episodes per suite).}
\label{tab:repro}
\begin{tabular}{lrrrrc}
\toprule
Suite & Published & Ours & $\Delta$ & 95\% Wilson CI & Gate ($\pm2$) \\
\midrule
LIBERO-Spatial & 97.6 & 97.8 & +0.2 & [96.1, 98.8] & \textbf{pass} \\
LIBERO-Object & 98.4 & 97.4 & -1.0 & [95.6, 98.5] & \textbf{pass} \\
LIBERO-Goal & 97.9 & 98.4 & +0.5 & [96.9, 99.2] & \textbf{pass} \\
LIBERO-Long (10) & 94.5 & 93.4 & -1.1 & [90.9, 95.3] & \textbf{pass} \\
\bottomrule
\end{tabular}

\end{widetable}

\begin{widetable}
\centering\small
\caption{Environment, checkpoint revisions and the pinned versions that the numbers in Tab.~\ref{tab:repro} were produced with.}
\label{tab:repro-env}
\begin{tabular}{ll}
\toprule
Component & Version / value \\
\midrule
\texttt{openvla-oft} & \texttt{SHA   e4287e94541f459edc4feabc4e181f537cd569a8} \\
\texttt{LIBERO} & \texttt{SHA        8f1084e3132a39270c3a13ebe37270a43ece2a01} \\
\texttt{python} & \texttt{Python 3.10.14} \\
\texttt{torch} & \texttt{2.2.0+cu121} \\
\texttt{MUJOCO\_GL} & \texttt{egl} \\
\texttt{gpu} & \texttt{NVIDIA A100 80GB PCIe, 81920 MiB} \\
\texttt{ckpt LIBERO-Spatial} & \texttt{moojink/openvla-7b-oft-finetuned-libero-spatial} \\
\texttt{ckpt LIBERO-Object} & \texttt{moojink/openvla-7b-oft-finetuned-libero-object} \\
\texttt{ckpt LIBERO-Goal} & \texttt{moojink/openvla-7b-oft-finetuned-libero-goal} \\
\texttt{ckpt LIBERO-Long (10)} & \texttt{moojink/openvla-7b-oft-finetuned-libero-10} \\
\bottomrule
\end{tabular}

\end{widetable}

Per-task rates are in Tab.~\ref{tab:repro-per-task}; the aggregate hides which tasks carry the residual failures. $\pi_{0.5}$ is run through openpi's server and its LIBERO client, unmodified except that the instruction is replaced, with the released \texttt{pi05\_libero} checkpoint. Its canonical success is \PiSpatialCanon, \PiObjectCanon, \PiGoalCanon{} and \PiLongCanon\% on Spatial, Object, Goal and Long ($n{=}250$, $100$, $250$, $100$) against the published $98.8$, $98.2$, $98.0$ and $92.4$~\citep{intelligence2025pi05}; each published value lies inside our Wilson $95\%$ interval.

\begin{widetable}
\centering\small
\caption{Per-task success rates for the reproduction runs.}
\label{tab:repro-per-task}
\resizebox{\linewidth}{!}{\begin{tabular}{llrr}
\toprule
Suite & Task & Succ.\ & $n$ \\
\midrule
LIBERO-Spatial & pick up the black bowl between the plate and the ramekin and place it on the plate & 100\% & 50 \\
LIBERO-Spatial & pick up the black bowl next to the ramekin and place it on the plate & 100\% & 50 \\
LIBERO-Spatial & pick up the black bowl from table center and place it on the plate & 100\% & 50 \\
LIBERO-Spatial & pick up the black bowl on the cookie box and place it on the plate & 98\% & 50 \\
LIBERO-Spatial & pick up the black bowl in the top drawer of the wooden cabinet and place it on the plate & 96\% & 50 \\
LIBERO-Spatial & pick up the black bowl on the ramekin and place it on the plate & 96\% & 50 \\
LIBERO-Spatial & pick up the black bowl next to the cookie box and place it on the plate & 100\% & 50 \\
LIBERO-Spatial & pick up the black bowl on the stove and place it on the plate & 92\% & 50 \\
LIBERO-Spatial & pick up the black bowl next to the plate and place it on the plate & 98\% & 50 \\
LIBERO-Spatial & pick up the black bowl on the wooden cabinet and place it on the plate & 98\% & 50 \\
\midrule
LIBERO-Object & pick up the alphabet soup and place it in the basket & 100\% & 50 \\
LIBERO-Object & pick up the cream cheese and place it in the basket & 98\% & 50 \\
LIBERO-Object & pick up the salad dressing and place it in the basket & 96\% & 50 \\
LIBERO-Object & pick up the bbq sauce and place it in the basket & 94\% & 50 \\
LIBERO-Object & pick up the ketchup and place it in the basket & 100\% & 50 \\
LIBERO-Object & pick up the tomato sauce and place it in the basket & 98\% & 50 \\
LIBERO-Object & pick up the butter and place it in the basket & 100\% & 50 \\
LIBERO-Object & pick up the milk and place it in the basket & 96\% & 50 \\
LIBERO-Object & pick up the chocolate pudding and place it in the basket & 94\% & 50 \\
LIBERO-Object & pick up the orange juice and place it in the basket & 98\% & 50 \\
\midrule
LIBERO-Goal & open the middle drawer of the cabinet & 98\% & 50 \\
LIBERO-Goal & put the bowl on the stove & 96\% & 50 \\
LIBERO-Goal & put the wine bottle on top of the cabinet & 100\% & 50 \\
LIBERO-Goal & open the top drawer and put the bowl inside & 94\% & 50 \\
LIBERO-Goal & put the bowl on top of the cabinet & 100\% & 50 \\
LIBERO-Goal & push the plate to the front of the stove & 98\% & 50 \\
LIBERO-Goal & put the cream cheese in the bowl & 100\% & 50 \\
LIBERO-Goal & turn on the stove & 100\% & 50 \\
LIBERO-Goal & put the bowl on the plate & 100\% & 50 \\
LIBERO-Goal & put the wine bottle on the rack & 98\% & 50 \\
\midrule
LIBERO-Long (10) & put both the alphabet soup and the tomato sauce in the basket & 90\% & 50 \\
LIBERO-Long (10) & put both the cream cheese box and the butter in the basket & 100\% & 50 \\
LIBERO-Long (10) & turn on the stove and put the moka pot on it & 96\% & 50 \\
LIBERO-Long (10) & put the black bowl in the bottom drawer of the cabinet and close it & 98\% & 50 \\
LIBERO-Long (10) & put the white mug on the left plate and put the yellow and white mug on the right plate & 94\% & 50 \\
LIBERO-Long (10) & pick up the book and place it in the back compartment of the caddy & 100\% & 50 \\
LIBERO-Long (10) & put the white mug on the plate and put the chocolate pudding to the right of the plate & 90\% & 50 \\
LIBERO-Long (10) & put both the alphabet soup and the cream cheese box in the basket & 98\% & 50 \\
LIBERO-Long (10) & put both moka pots on the stove & 74\% & 50 \\
LIBERO-Long (10) & put the yellow and white mug in the microwave and close it & 94\% & 50 \\
\bottomrule
\end{tabular}
}
\end{widetable}

\paragraph{Controller-routing bug.} Until the harness was audited, \texttt{build\_controller} matched controller names by prefix, so the name \texttt{canonical-unless-outlier} was routed to the plain canonical controller. Every row under that name in the earlier grids is therefore a second run of the baseline; those rows are relabelled ``canonical (replicate run)'' in the tables and are what we use as the run-to-run noise floor (\S\ref{sec:setup}). Manifests now record the controller class actually instantiated, a regression test covers the routing, and the real controller was run afresh (App.~\ref{app:horizon}).

\subsection{A library fork that silently removes causal masking}
\label{app:causal}
The weak-policy arm of \S\ref{sec:weak} required a second reproduction, and it failed the first time in a way worth reporting: base OpenVLA scored $30.0\%$ on LIBERO-Spatial against a published $84.7\%$, with no error, no warning, and actions that looked entirely plausible.

The cause is not in the policy, the checkpoint, or the preprocessing. Three hypotheses were tested by forward pass and all three were wrong: the checkpoint's own remote modeling code is what loads (not the OFT classes registered into the same \texttt{Auto*} namespace); the attention mask that the checkpoint's \texttt{predict\_action} leaves one token short of \texttt{input\_ids} changes nothing; and language does reach the model, since swapping the instruction moves the predicted action by $0.20$--$0.66$ normalized units. The cause was found by diffing the environment. OpenVLA-OFT requires a fork of \texttt{transformers}, which reports itself as version 4.40.1 and differs from that release in exactly one file and one hunk: in \texttt{LlamaSdpaAttention.forward} it passes \texttt{is\_causal=False} and overwrites the causal mask with its own last row. OFT needs this: its $H \times D$ action tokens are decoded in a single forward pass and must attend to each other, but for an un-padded single sequence stock \texttt{transformers} leaves the causal mask \texttt{None} and relies on the \texttt{is\_causal} flag, so under the fork \emph{there is no causal mask at all} and the prompt is encoded bidirectionally.

Whether that is fatal depends on the action head, which is why it survived undetected in an OFT-only stack. OFT reads one $L_1$ regression head off the final hidden states and reproduces its published numbers on the fork (Tab.~\ref{tab:repro}). An autoregressive policy takes $D$ sequential $\arg\max$ decisions over 256 action bins and feeds each back in, so the same perturbation compounds: on byte-identical pixels and the same instruction, the fork and stock 4.40.1 choose different action bins in three of seven dimensions. Restoring causal attention, with a stock \texttt{transformers} installed beside the fork and placed first on \texttt{sys.path}, since eager attention instead dies on the mask off-by-one that the SDPA path silently discards, recovers the published behavior (Tab.~\ref{tab:weak-repro}).

\begin{widetable}
\centering\small
\caption{Base OpenVLA through our harness, before and after restoring causal attention, against the published reference (OpenVLA's own README; 500 trials per suite). Intervals are Wilson 95\% intervals on our rate; $\dagger$: the interval does not reach the published point estimate. LIBERO-Long appears twice because two independent runs of the same configuration exist: $50.0$ is the paired reference of \S\ref{sec:weak}, $55.0$ that of Tab.~\ref{tab:probe}.}
\label{tab:weak-repro}
\resizebox{\linewidth}{!}{\begin{tabular}{llrrlrl}
\toprule
Suite & Controller & $K$ & $n$ & Success (95\% CI) & Published & transformers \\
\midrule
LIBERO-Long (10) & \texttt{canonical-1} & 1 & 100 & $50.0$ $[40.4, 59.6]$ & $53.7 \pm 1.3$ & stock 4.40.1 \\
LIBERO-Long (10) & \texttt{canonical-1} & 1 & 100 & $55.0$ $[45.2, 64.4]$ & $53.7 \pm 1.3$ & stock 4.40.1 \\
LIBERO-Goal & \texttt{canonical-1} & 1 & 250 & $76.0$ $[70.3, 80.9]$ & $79.2 \pm 1.0$ & stock 4.40.1 \\
LIBERO-Object & \texttt{canonical-1} & 1 & 250 & $87.6$ $[82.9, 91.1]$ & $88.4 \pm 0.8$ & stock 4.40.1 \\
LIBERO-Spatial & \texttt{canonical-1} & 1 & 20 & $30.0$ $[14.5, 51.9]$$^{\dagger}$ & $84.7 \pm 0.9$ & OFT fork (non-causal) \\
LIBERO-Spatial & \texttt{canonical-1} & 1 & 250 & $85.6$ $[80.7, 89.4]$ & $84.7 \pm 0.9$ & stock 4.40.1 \\
\bottomrule
\end{tabular}
}
\end{widetable}

We report this at length for two reasons. It is a reproducibility hazard that generalizes: any autoregressive model evaluated inside an OpenVLA-OFT environment is affected, silently. And it is the reason our own guard is behavioral rather than a version check: the fork and the release carry the same version string, so the adapter builds a two-layer Llama at load time and refuses to run if position $0$'s hidden state moves when only later tokens change. A degraded baseline is the most expensive kind of bug in a paper whose every claim is a difference against one.

\section{The Strong Policy: Every Controller, Both Suites}
\label{app:std}
\begin{widetable}
\centering\small
\caption{Shifted LIBERO-Spatial (distractor mug), OpenVLA-OFT, $n{=}500$ per arm paired on identical initial states. Generated by \texttt{lcc/analysis/main\_table.py}.}
\label{tab:main-shift}
\begin{tabular}{lrrlr}
\toprule
Method & $n$ & Succ.\ (\%) & $\Delta$ vs.\ ref.\ [95\% CI] & $p_{\text{Holm}}$ \\
\midrule
\multicolumn{5}{l}{\textit{OpenVLA-OFT / LIBERO-Spatial with distractor, reference: Canonical-8}} \\
\midrule
Canonical-8 & 500 & 88.6 \tiny[85.5, 91.1] & --- & --- \\
Paraphrase medoid, $K{=}4$ & 500 & 90.0 \tiny[87.1, 92.3] & +1.4 [-0.6, 3.4] & 0.743 \\
Paraphrase mean, $K{=}4$ & 500 & 87.4 \tiny[84.2, 90.0] & -1.2 [-3.8, 1.2] & 0.859 \\
One of $K{=}4$ instructions (canonical incl.), per episode & 500 & 87.8 \tiny[84.6, 90.4] & -0.8 [-3.2, 1.6] & 0.859 \\
\bottomrule
\end{tabular}

\end{widetable}
All controllers on standard LIBERO-Spatial, OpenVLA-OFT, paired on identical initial states; $n$ per row, with three rows from runs that stopped short of $500$. Nothing is significant after Holm correction, and \S\ref{sec:setup} explains why no improvement could be: the headroom is at or below the minimum detectable effect. The adaptive rows carry the calibration source in brackets because thresholds calibrated on another suite are not valid (App.~\ref{app:horizon}).

\begin{widetable}
\centering\small
\caption{Standard LIBERO-Spatial, all controllers. Generated by \texttt{lcc/analysis/main\_table.py}.}
\label{tab:main-std}
\resizebox{\linewidth}{!}{\begin{tabular}{lrrlr}
\toprule
Method & $n$ & Succ.\ (\%) & $\Delta$ vs.\ ref.\ [95\% CI] & $p_{\text{Holm}}$ \\
\midrule
\multicolumn{5}{l}{\textit{OpenVLA-OFT / LIBERO-Spatial, reference: Canonical-8}} \\
\midrule
Canonical-8 & 500 & 97.2 \tiny[95.4, 98.3] & --- & --- \\
Canonical (replicate run) & 500 & 98.0 \tiny[96.4, 98.9] & +0.8 [0.2, 1.6] & 1.000 \\
Canonical (greedy) & 408 & 95.6 \tiny[93.1, 97.2] & -1.0 [-3.4, 1.5] & 1.000 \\
Canonical-4 & 430 & 98.1 \tiny[96.4, 99.1] & +0.5 [-1.2, 2.1] & 1.000 \\
Gated LCC & 500 & 97.2 \tiny[95.4, 98.3] & +0.0 [-1.4, 1.4] & 1.000 \\
Paraphrase medoid, $K{=}4$ & 500 & 96.0 \tiny[93.9, 97.4] & -1.2 [-2.8, 0.4] & 1.000 \\
Paraphrase mean, $K{=}4$ & 405 & 97.0 \tiny[94.9, 98.3] & +0.2 [-1.2, 2.0] & 1.000 \\
Monitor-only & 500 & 97.0 \tiny[95.1, 98.2] & -0.2 [-1.2, 0.8] & 1.000 \\
One paraphrase, fixed per episode & 500 & 96.2 \tiny[94.1, 97.6] & -1.0 [-3.0, 1.0] & 1.000 \\
One of $K{=}4$ instructions (canonical incl.), per episode & 500 & 96.4 \tiny[94.4, 97.7] & -0.8 [-2.4, 0.8] & 1.000 \\
One of $K{=}4$ instructions (canonical incl.), per replan & 500 & 96.8 \tiny[94.9, 98.0] & -0.4 [-2.0, 1.2] & 1.000 \\
Training instruction only & 500 & 97.4 \tiny[95.6, 98.5] & +0.2 [-0.6, 1.0] & 1.000 \\
Paraphrase 1 only & 500 & 95.8 \tiny[93.7, 97.2] & -1.4 [-3.4, 0.6] & 1.000 \\
Paraphrase 2 only & 500 & 95.6 \tiny[93.4, 97.1] & -1.6 [-3.4, 0.2] & 1.000 \\
Paraphrase 3 only & 500 & 96.2 \tiny[94.1, 97.6] & -1.0 [-2.8, 0.6] & 1.000 \\
\bottomrule
\end{tabular}
}
\end{widetable}

\section{Latency and Compute}
\label{app:cost}
\begin{widetable}
\centering\small
\caption{Measured wall-clock per replanning step for OpenVLA-OFT on one idle A100-40\,GB (no other compute on the card), median and p95 of 20 CUDA-synchronised repetitions. \emph{Sequential} is $K$ separate reference calls; \emph{batched} computes the vision features once and issues one language-model forward per distinct prompt token length. Peak GPU allocation was 14.98\,GiB at $K{=}1$ and 16.66\,GiB at $K{=}8$. }
\label{tab:cost}
\begin{tabular}{rlrrrrr}
\toprule
$K$ & mode & median (ms) & p95 (ms) & $\times$ vs.\ $K{=}1$ & control rate (Hz, $H{=}8$) & peak GiB \\
\midrule
1 & sequential & 101 & 104 & 1.00 & 79.3 & 15.0 \\
2 & batched & 190 & 193 & 1.89 & 42.0 & 15.0 \\
2 & sequential & 201 & 202 & 1.99 & 39.8 & 15.0 \\
4 & batched & 301 & 303 & 2.98 & 26.6 & 15.5 \\
4 & sequential & 406 & 409 & 4.02 & 19.7 & 15.0 \\
8 & batched & 583 & 586 & 5.78 & 13.7 & 16.7 \\
8 & sequential & 814 & 826 & 8.06 & 9.8 & 15.0 \\
\bottomrule
\end{tabular}

\end{widetable}
Tab.~\ref{tab:cost} gives the measurement behind the $2.98\times$ figure of \S\ref{sec:method}: wall-clock per replanning step by $K$ and batching mode, CUDA-synchronised, on an idle card. Sequential cost is exactly linear in $K$; the grouped implementation removes only the fixed per-call overhead, because the transformer forward is compute-bound at these sequence lengths.

\section{Runtime Failure Detection and the Length Confound}
\label{app:failure}
\paragraph{Episode length leaks the label.}
The natural episode score ($\max_t \bar U_t$ over a sliding-window mean of $\ut$) cannot be used, and neither can any other statistic of a whole trace. On a benchmark with early termination a failed episode runs to the step limit while a successful one stops when the success predicate fires: on LIBERO-Spatial our successes take a median of 13 replanning steps and our failures 28, and \textbf{replan count alone predicts failure with AUROC 1.000}. A max over a longer trace is larger for that reason alone, so a whole-trace score inherits near-perfect separation while carrying no information about the quantity being scored.

The artefact is not small. Scored over the whole trace, the 500 standard-Spatial episodes give AUROC 0.836 for language disagreement, 0.959 for visual TTA, 1.000 for activation perturbation and 1.000 for the free temporal-consistency monitor. Restricted to a fixed prefix of five replanning steps, which every episode reaches, so none is dropped and no length information survives: the same detectors give 0.480, 0.444, 0.539 and 0.629, every interval containing $0.5$. We report both, because the gap between them is the measurement that matters here. Whether published detectors share this confound depends on their scoring protocol, which we have not audited; the construction is generic to any early-terminating benchmark, and ours had it.

\paragraph{Risk-set detection.}
A short fixed prefix removes the confound but can only ask about \emph{early} detection. To ask about late detection we borrow the risk-set construction from survival analysis: at replanning step $t$, restrict attention to the episodes \emph{still running} at $t$ and ask how well disagreement observed up to $t$ separates eventual failures from eventual successes. Every episode in that comparison has survived equally long, so length is uninformative, and $t$ may be late. The result is an AUROC curve over $t$  rather than a single number, reported with the risk-set size at each $t$ so the reader can see where the estimate thins. Conditioning on survival is itself a selection: late risk sets contain the slow successes rather than a random sample, and we state that rather than treating a late AUROC as unconditional; it is nonetheless exactly the question a runtime monitor faces.

We report risk-set AUROC, AUPRC, and false-alarm rate at 80\% recall. Comparators, all scored from the same shadow rollouts on identical trajectories: visual test-time augmentation, activation perturbation (Gaussian noise on hidden activations, in the spirit of PFD~\citep{lee2026pfd}), and action-chunk temporal consistency~\citep{seligmann2026vlafail}, together with episode length itself as the baseline every detector must beat. Sampling variance does not exist for OFT's deterministic head; STAC~\citep{agia2024stac}, token entropy~\citep{karli2025askbefore} and the supervised SAFE~\citep{gu2025safe} were not run.

\paragraph{Result: language disagreement is the only signal informative from the first replan.}
On standard LIBERO-Spatial the shadow rollouts have 14 failures in 500 episodes and every length-controlled interval contains $0.5$; that is an absence of power, not a null, and we report it as such. The measurement is therefore made in the shift regime, on a separate canonical run of the same configuration ($500$ episodes, $51$ failures, i.e.\ $89.8\%$, within the run-to-run noise of the $88.6\%$ reference of Tab.~\ref{tab:main-shift}).

There, for $t=1$ through $t=7$, paraphrase disagreement is significant at every replanning step (risk-set AUROC $0.598$--$0.620$, 95\% intervals excluding $0.5$) and \emph{every other source is at chance}, including the free temporal-consistency monitor, and including visual test-time augmentation, which carries four times the disagreement magnitude (App.~\ref{sec:sensitivity}). From $t\approx 8$ the others become informative and by $t\approx 12$ activation perturbation and temporal consistency overtake it, reaching $0.70$--$0.79$.

The ordering has a mechanism rather than being a coincidence of this split. Temporal consistency compares the previously executed chunk against the fresh one, so it does not exist at $t=1$ at all and needs several replans before its history carries information; the visual and activation ensembles need the episode to reach a state in which their perturbation bites. Paraphrase disagreement is available at the first query and informative immediately. For a runtime monitor that is the distinction that matters: a detection at $t=1$ leaves the entire episode in which to intervene, and one at $t=12$ leaves very little.

We state the limits plainly. The intervals are per detector and per step, uncorrected for the four sources and seven steps compared; an AUROC of $0.60$ is a small early signal, not a strong detector. The late-$t$ figures for every source rest on risk sets thinned to 85--175 episodes composed disproportionately of slow successes. And the claim is about \emph{when} a signal becomes available, not about beating a supervised detector such as SAFE~\citep{gu2025safe}. This is a measurement on one checkpoint and one shift; we do not state it as a general property.

\begin{widetable}
\centering\small
\caption{Length-controlled failure detection on the shifted suite (OpenVLA-OFT, $500$ shadow episodes, $51$ failures), every detector scored on the same rollouts over a fixed prefix of five replanning steps; the first row, replan count scored on the whole trace, is the confound. AUROC@2: the same statistic over a prefix of two replans; FAR@80\%rec: false-alarm rate at 80\% recall; ACC: action-chunk consistency. Generated by \texttt{lcc/analysis/failure.py}.}
\label{tab:failure}
\begin{tabular}{lcccc}
\toprule
Detector & AUROC [95\% CI] & AUROC@2 & AUPRC & FAR@80\%rec \\
\midrule
Replan count (length only) & 0.999 [0.997, 1.000] & 0.999 & 0.981 & 0.002 \\
Language (ours) & 0.600 [0.512, 0.683] & 0.617 & 0.179 & 0.675 \\
Visual TTA & 0.512 [0.436, 0.590] & 0.522 & 0.102 & 0.641 \\
Activation pert. & 0.509 [0.428, 0.589] & 0.534 & 0.103 & 0.773 \\
Temporal consistency (ACC) & 0.524 [0.430, 0.622] & 0.454 & 0.121 & 0.842 \\
\bottomrule
\end{tabular}

\end{widetable}

\begin{widefigure}
\centering
\includegraphics[width=0.9\linewidth]{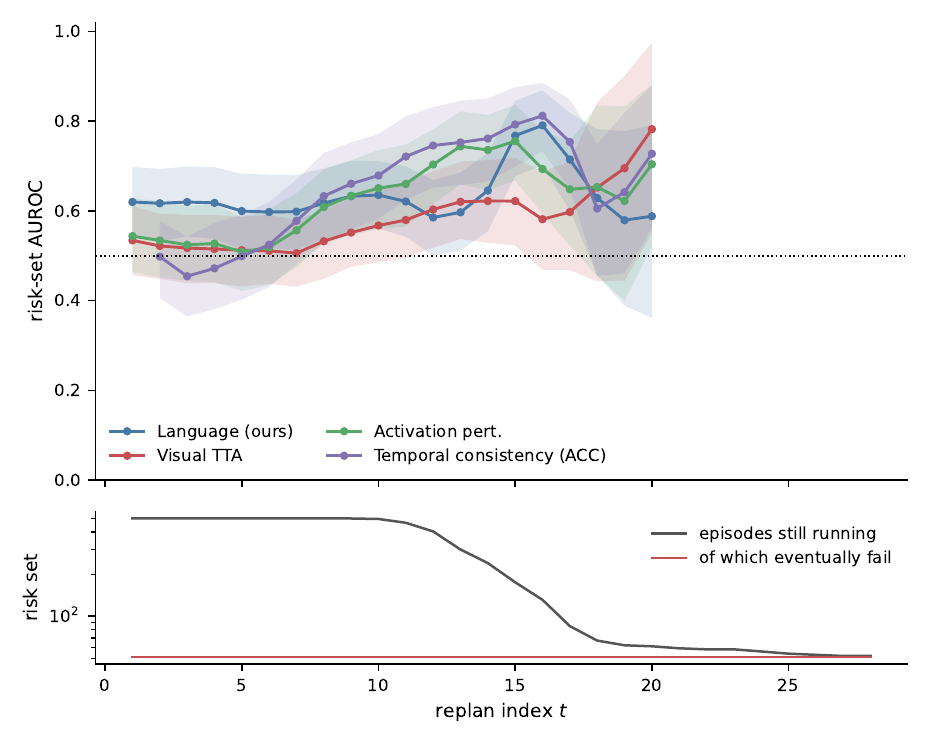}
\caption{Risk-set AUROC by replanning step on the shifted suite, one curve per ensemble source; the risk set at $t$ is the episodes still running at $t$, so length cannot leak. Generated by \texttt{lcc/analysis/riskset.py}.}
\label{fig:riskset}
\end{widefigure}

\section{How Paraphrase-Sensitive Are VLAs?}
\label{sec:sensitivity}
\paragraph{Protocol.}
For each model and benchmark split we roll out the canonical controller and, at every replanning step, additionally compute (without executing) the $K$-member ensembles from four sources at matched $K$: language paraphrases (ours), stochastic action sampling (flow/diffusion policies only; different initial noise, same instruction), visual test-time augmentation (random-offset crops and mild photometric jitter), and hidden-activation perturbation following~\citet{lee2026pfd}. Each yields a disagreement trace via Eq.~\eqref{eq:u}.

Because the executed action is always the canonical one, the executed trajectory is identical to a plain canonical rollout: disagreement is measured, never acted on. Every source is therefore compared \emph{on the same states}, which matters more than it might appear: separate rollouts per source would leave any difference between their disagreement distributions attributable to the states each controller happened to visit rather than to the perturbation. One set of rollouts consequently yields this section and all of App.~\ref{app:failure} for every source. Two implementation facts are worth stating, because they bound what the comparison can mean. First, sources are not equally cheap: only the language ensemble shares its observation and can be grouped into fewer forward passes, while visual augmentation changes the image and activation perturbation changes the weights, so both cost $K$ forwards; we report that asymmetry rather than amortising it. Second, per-member perturbations are drawn once per episode and held fixed, so a member is a consistent ``point of view'' rather than fresh noise at each step, since redrawing would make the measured disagreement largely a measure of resampling.

A source that cannot apply to a policy is reported as \emph{not applicable}, with the reason, and never as a zero: the OpenVLA-OFT checkpoints use an L1-regression head, so a sampling ensemble over them is identically zero, and a blank rendered as $0$ would later read as ``sampling produces little disagreement''.

\paragraph{Result: magnitude and informativeness are decoupled.}
Tab.~\ref{tab:sens} and Fig.~\ref{fig:sensitivity} give per-replan disagreement by source on OpenVLA-OFT on the shifted LIBERO-Spatial suite of App.~\ref{app:failure} ($7{,}429$ replanning steps over $500$ shadow episodes). Visual test-time augmentation produces about four times the action-level disagreement of paraphrase and activation perturbation about half; sampling is not applicable to a deterministic regression head and is reported as such rather than as zero. Every source is larger on failed episodes than on successful ones, by similar ratios (roughly $1.5\times$ at the median). Yet under the length-controlled protocol of App.~\ref{app:failure} the largest perturbation is uninformative about failure at every horizon while the smallest, paraphrase, is the only one significant from the first replanning step: magnitude and informativeness are decoupled, and a study reporting magnitudes alone, which is what this section was originally planned to be, would have ranked the sources in the wrong order. At the success level, executing one paraphrase instead of the training instruction changes OpenVLA-OFT's success by $\OftSpatialParaDelta$, $\OftObjectParaDelta$ and $\OftLongParaDelta$ points on Spatial, Object and Long, each within noise, and by $\OftGoalParaDelta$ on Goal (Tab.~\ref{tab:reliance}). On the first three suites the policy hears the rewording, acts measurably differently, and almost always recovers.

\begin{widefigure}
\centering
\includegraphics[width=0.95\linewidth]{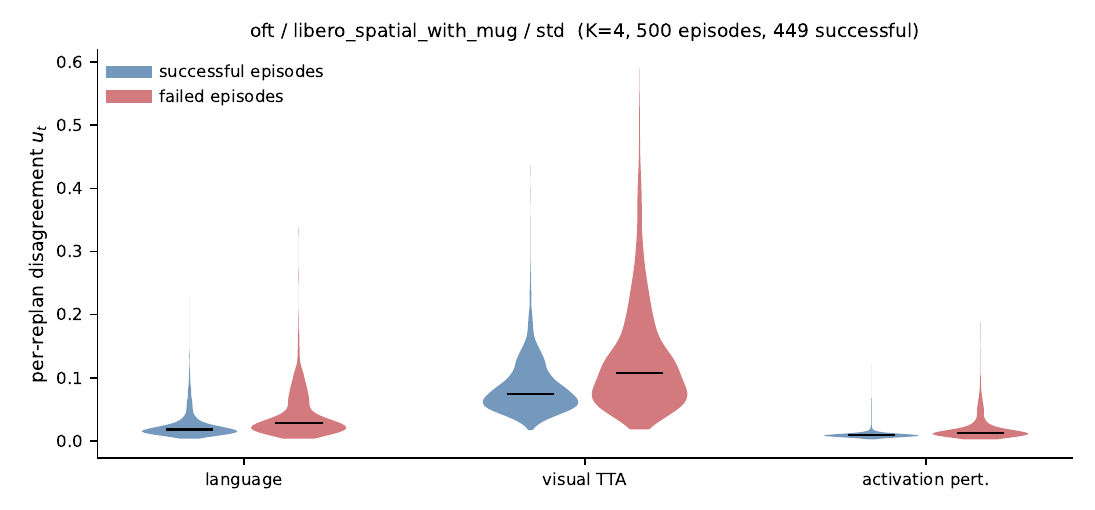}
\caption{Per-replan disagreement $u_t$ by ensemble source on identical canonical trajectories (OpenVLA-OFT, shifted LIBERO-Spatial). Generated by \texttt{lcc/analysis/sensitivity.py}.}
\label{fig:sensitivity}
\end{widefigure}

\begin{widetable}
\centering\small
\caption{Action-level disagreement by source and by episode outcome (OpenVLA-OFT, shifted LIBERO-Spatial, $500$ shadow episodes). Generated by \texttt{lcc/analysis/sensitivity.py}.}
\label{tab:sens}
\resizebox{\linewidth}{!}{\begin{tabular}{llrrrr}
\toprule
\multicolumn{6}{l}{\textbf{Action-level disagreement by ensemble source}} \\
Source & & median & mean & p90 & $n$ replans \\
\midrule
language ($\times$1.00) & & 0.0191 & 0.0275 & 0.0599 & 7429 \\
visual TTA ($\times$4.09) & & 0.0782 & 0.0932 & 0.1527 & 7429 \\
activation pert. ($\times$0.48) & & 0.0091 & 0.0119 & 0.0167 & 7429 \\
sampling & \multicolumn{5}{p{0.72\linewidth}}{n/a --- policy has a deterministic action head (L1 regression), so a sampling ensemble is identically zero; report the cell as n/a} \\
\midrule
language, successful episodes & & 0.0179 & 0.0243 & 0.0486 & 6001 \\
language, failed episodes & & 0.0277 & 0.0410 & 0.0883 & 1428 \\
\midrule
visual TTA, successful episodes & & 0.0748 & 0.0845 & 0.1372 & 6001 \\
visual TTA, failed episodes & & 0.1080 & 0.1299 & 0.2395 & 1428 \\
\midrule
activation pert., successful episodes & & 0.0087 & 0.0100 & 0.0132 & 6001 \\
activation pert., failed episodes & & 0.0128 & 0.0200 & 0.0462 & 1428 \\
\bottomrule
\end{tabular}
}
\end{widetable}

\section{Adaptive Horizon, Conformal Calibration, and Deployment Modes}
\label{app:horizon}
These components were specified before the control results existed and are retained for completeness: the adaptive horizon and escalation rule, its conformal calibration, and the gated / monitor / distilled deployment modes. None survives the results of \S\ref{sec:weak} as a deployment recommendation; the distillation target in particular (the executed consensus action) is now known to be worse than the canonical action, and the training-time remedy of \S\ref{sec:aug} replaces it.

\subsection{Adaptive horizon and escalation}
\label{sec:horizon}
The execution horizon $h_t$ is set from $\ut$:
\begin{equation}
h_t=\begin{cases}
H, & \ut<\tau_{\text{low}},\\
\max(1,\lfloor H/2\rfloor), & \tau_{\text{low}}\le \ut<\tau_{\text{high}},\\
1, & \ut\ge\tau_{\text{high}},
\end{cases}
\label{eq:horizon}
\end{equation}
so the controller runs open-loop for longer when the paraphrases agree and re-observes after fewer actions when they do not. To prevent chattering, $h_t$ may only increase after $m$ consecutive replans below the relevant threshold (hysteresis). A continuous variant, $h_t=\mathrm{clip}\bigl(\mathrm{round}(H e^{-(\ut-\tau_{\text{low}})/\kappa}),1,H\bigr)$ for $\ut\ge\tau_{\text{low}}$, is ablated.

\paragraph{Calibration without leakage.}
Thresholds are set on a calibration split disjoint from evaluation: either another suite (LIBERO-Object, for evaluation on Spatial) or held-out tasks within the suite (tasks 0--4 for evaluation on tasks 5--9), and we report cross-suite transfer. $\tau_{\text{low}}$ is a percentile of calibration disagreement as in the calibration run. $\tau_{\text{high}}$ is set by a conformal false-alarm rule: let $\{u^{(i)}\}_{i=1}^{n}$ be the per-replan disagreements from \emph{successful} calibration rollouts; choosing $\tau_{\text{high}}$ as their $\lceil(1-\alpha)(n+1)\rceil$-th order statistic yields $\Pr(\ut>\tau_{\text{high}}\mid\text{success})\le\alpha$ for a new exchangeable replan~\citep{angelopoulos2021gentle}. We report $\alpha\in\{0.05,0.1,0.2\}$.

\paragraph{Where that guarantee holds, and where it does not.}
Exchangeability is a real precondition and it fails in two ways we measured rather than assumed. First, \emph{across suites}: LIBERO-Spatial's per-replan disagreement is about twice LIBERO-Object's (medians $0.023$ vs $0.012$), so thresholds calibrated on one suite and applied to the other put $39\%$ of replans above a $\tau_{\text{high}}$ built for $10\%$ and escalate on half of all episodes. Calibration must be in-suite; since LIBERO ships 50 initial states per task, that costs either held-out tasks or held-out initial states, and we take the held-out-task split.

Second, and less obviously, \emph{within} a suite the adaptive controller's own replans are not exchangeable with the calibration pool. Calibration draws from canonical rollouts, whose replan times are fixed; the adaptive controller replans \emph{more often when $\ut$ is high}, which is its mechanism, so its replan stream over-samples exactly the states being thresholded. With in-suite calibration at $\alpha=0.1$ we measure an exceedance rate of $15.6\%$ on canonical rollouts and $33.3\%$ on the adaptive controller's own stream, and the inflation appears within every individual task, so it is selection rather than a different mix of tasks.

We therefore claim the conformal false-alarm rate for \textbf{monitor-only mode}, whose ensemble fires on a schedule independent of $\ut$ and is exchangeable with the calibration pool, and for the adaptive controller we report the \emph{measured} exceedance rate beside the nominal $\alpha$. A valid guarantee for the adaptive case needs either a fixed point (calibrating on the controller's own stream and iterating) or a covariate-shift-weighted conformal treatment. We leave both to future work and do not claim the guarantee for the plain rule.

\paragraph{Escalation.}
If $\ut\ge\tau_{\text{high}}$ for $m$ consecutive replans the controller escalates. In simulation an escalation is logged as a detection event and the episode continues, since no recovery behaviour exists for these benchmarks; the detection results are in App.~\ref{app:failure}. Algorithm~\ref{alg:lcc} summarizes one replanning step.

\begin{algorithm}[t]
\caption{\method: one replanning step}
\label{alg:lcc}
\begin{algorithmic}[1]
\Require frozen $\pi$, instruction set $\mathcal{L}$, scale $s$, thresholds $\tau_{\text{low}},\tau_{\text{high}}$, chunk length $H$, escalation patience $T$
\State $\{A^{(k)}_t\}_{k=1}^K \gets \pi(o_t,\mathcal{L})$ \Comment{vision features computed once; language forwards grouped by prompt length (App.~\ref{app:impl})}
\State compute $d(A^{(i)}_t,A^{(j)}_t)$ for all $i<j$ via Eq.~\eqref{eq:u}; $\ut$ via Eq.~\eqref{eq:u}
\State $A^\star_t \gets$ medoid via \S\ref{sec:medoid}
\State $h_t \gets$ Eq.~\eqref{eq:horizon} with hysteresis
\If{$\ut\ge\tau_{\text{high}}$ for $T$ consecutive steps} \State escalate (recovery / ask for help) \EndIf
\State execute the first $h_t$ actions of $A^\star_t$; re-observe
\end{algorithmic}
\end{algorithm}

\subsection{Deployment modes}
\label{sec:deploy-method}
Three deployment modes were specified around full \method{} as the teacher; \S\ref{sec:deploy} states which were run.

\paragraph{Gated \method.}
Receding-horizon execution gives a free consistency signal: the previously executed chunk and the newly predicted canonical chunk overlap in time~\citep{seligmann2026vlafail}. Let $c_t$ be the normalized distance (Eq.~\ref{eq:u}) between the overlapping segment of $A^\star_{t-1}$, shifted by $h_{t-1}$, and the corresponding steps of the fresh canonical chunk $\pi(o_t,\ell_1)$. If $c_t<\tau_{\text{gate}}$ the canonical chunk is executed with horizon $H$; otherwise the $K$-query consensus and adaptive horizon of Algorithm~\ref{alg:lcc} run for this step. $\tau_{\text{gate}}$ is calibrated on the disjoint split, and sweeping it traces a curve from the canonical controller (never fire) to full \method{} (always fire). We report trigger rate and mean queries per replan.

One detail is easy to overlook and changes what can be claimed: that overlap has length $H-h_{t-1}$, so at $h=H$ (the default for these checkpoints, and exactly what Canonical-8 is), consecutive chunks are \emph{disjoint in time} and $c_t$ is undefined at every replan. A gate defined only on the overlap therefore never fires in the configuration the main table uses. We keep the signal free by defining two regimes and recording which one produced every logged value: for $h<H$, the overlap distance above; for $h=H$, the distance between the last action actually executed and the first action of the new chunk, which is an equally free measure of whether the policy revised its plan across the replan boundary. The two are \emph{not} interchangeable: one step versus $H-h$ steps, and their medians differed by roughly $40\%$ on a held-out episode, so a $\tau_{\text{gate}}$ calibrated in one regime is not valid in the other; the cost--success sweep is reported within a regime rather than across the two. Neither variant exists on an episode's first replan, so the gate cannot fire there and mean queries per replan is computed over all replans including it.

\paragraph{Monitor-only.}
The ensemble runs every $N$-th replan and $\ut$ is used solely for escalation (\S\ref{sec:horizon}); the executed action is canonical. This isolates the failure-detection value at a chosen budget.

\paragraph{Distilled student.}
Let $\mathcal{D}_{\method}=\{(o,\ell_k,A^\star)\}$ collect, from full-\method{} rollouts in the shift regimes, every observation together with the executed medoid $A^\star$ and \emph{each} of the $K$ instructions, restricted to successful episodes (success labels are free in simulation) or, without labels, to episodes with low mean disagreement. A student $\pi_\theta$ initialized from the teacher is fine-tuned with
\begin{align}
\mathcal{L}(\theta)&=\underbrace{\mathbb{E}_{\mathcal{D}_{\method}}\bigl\lVert\pi_\theta(o,\ell_k)-A^\star\bigr\rVert_1}_{\text{consensus distillation}} \label{eq:uill}\\
&\quad+\lambda_{\text{cons}}\underbrace{\mathbb{E}_{\mathcal{D}_{\method}}\,d\bigl(\pi_\theta(o,\ell_i),\pi_\theta(o,\ell_j)\bigr)}_{\text{paraphrase consistency}}\nonumber\\
&\quad+\underbrace{\mathcal{L}_{\text{BC}}(\theta;\mathcal{D}_{\text{train}})}_{\text{replay}},\nonumber
\end{align}
where the replay term is the original behavior-cloning loss on the original multi-task data. Replay is not a regularizer for forgetting only: a student trained to be invariant to \emph{paraphrase} can become invariant to \emph{language}, the failure mode documented by~\citet{fei2025liberoplus}, and the original data, which pairs different instructions with different behaviors in the same scenes, is what keeps it grounded. The wrong-instruction probe of \S\ref{sec:probe} is the guard against that failure mode. The student is a single-query controller with the teacher's cost. Fine-tuning uses the LoRA recipe of each policy family (App.~\ref{app:post}).

\subsection{Deployment modes as measured}
\label{sec:deploy}
Of the three modes, only monitor-only was run to completion (standard LIBERO-Spatial, $N{=}4$: $-0.2$ $[-1.2,+0.8]$, Tab.~\ref{tab:main-std}), together with the gated variant at one operating point ($+0.0$ $[-1.4,+1.4]$). Neither the $\tau_{\text{gate}}$ sweep nor the distilled student was pursued once the results of \S\ref{sec:weak} were in: the gate's trigger signal predicts failure only from about the eighth replanning step once episode length is controlled (App.~\ref{app:failure}), and the distillation target, the executed consensus action, is measured to be worse than the canonical action, so a student trained on it would be trained toward a worse policy. The training-time remedy of \S\ref{sec:aug} replaces it.

\paragraph{Canonical-unless-outlier, measured.} The real controller (execute the canonical action unless it is the outlier among the $K{=}4$ members (total distance more than two robust deviations above the median), then the medoid) scores $83.2\%$ on base OpenVLA / LIBERO-Spatial: $-2.4$ $[-7.6,+2.8]$, not significant, with discordance $0.18$ against canonical, so it is a genuinely different controller (unlike the mis-routed rows) and roughly neutral. Keeping the privileged member avoids most of the harm and gains nothing. Tab.~\ref{tab:ksweep} places it beside the $K$ sweep.

\begin{widetable}
\centering\small
\caption{Paraphrase consensus at $K\in\{2,4,8\}$ and the outlier-gated variant, base OpenVLA / LIBERO-Spatial, paired against canonical greedy. $K{=}2$ tie-breaks to the canonical member by construction. Generated by \texttt{lcc/analysis/main\_table.py}.}
\label{tab:ksweep}
\begin{tabular}{lrrlr}
\toprule
Method & $n$ & Succ.\ (\%) & $\Delta$ vs.\ ref.\ [95\% CI] & $p_{\text{Holm}}$ \\
\midrule
\multicolumn{5}{l}{\textit{base OpenVLA / LIBERO-Spatial, reference: Canonical (greedy)}} \\
\midrule
Canonical (greedy) & 250 & 85.6 \tiny[80.7, 89.4] & --- & --- \\
Canonical unless outlier & 250 & 83.2 \tiny[78.1, 87.3] & -2.4 [-7.6, 2.8] & 0.903 \\
Paraphrase medoid [K=2] & 250 & 85.6 \tiny[80.7, 89.4] & +0.0 [-3.6, 3.6] & 1.000 \\
Paraphrase medoid [K=4] & 250 & 68.8 \tiny[62.8, 74.2] & \textbf{-16.8 [-22.8, -10.8]} & $<$0.001 \\
Paraphrase medoid [K=8] & 100 & 74.0 \tiny[64.6, 81.6] & -10.0 [-21.0, 1.0] & 0.331 \\
\bottomrule
\end{tabular}

\end{widetable}

\section{Human Equivalence Check of the Paraphrases}
\label{app:human}
The paraphrase pipeline accepts a candidate when an LLM judge rates it equivalent and its embedding cosine clears a floor (App.~\ref{app:paraphrases}). To check that filter, one author labelled every paraphrase that an evaluation run executed (\HumanN{} pairs: three per task for each type on Spatial and for the \emph{llm} set on the other suites) as the same task or a different one, with the type and suite hidden and the order shuffled; the criterion was that a robot doing exactly the training instruction would also have done exactly the paraphrase. \HumanDiff{} pairs (\HumanDiffPct\%) were judged to change the task; no reason was recorded per pair. Tab.~\ref{tab:human} recomputes every one-paraphrase cost on the episodes whose executed paraphrase was judged equivalent, paired with the same canonical episodes. Every rescored value lies inside the original $95\%$ interval. Read the other way, as a contrast set~\citep{gardner2020contrast}, the pairs judged task-changing cost more where the instruction is the task: on Goal $\BaseGoalHumanDiff$ for base OpenVLA and $\OftGoalHumanDiff$ for OFT, against $\BaseGoalHuman$ and $\OftGoalHuman$ on the equivalent pairs, so the policies do follow those changes. On Spatial the augmented model's cost is $\AugLlmHumanDiff$ on its task-changing pairs and $\AugLlmHuman$ on the rest, a smaller gap than the control's ($\CtrlLlmHumanDiff$ against $\CtrlLlmHuman$); with one annotator's labels this cannot separate over-invariance from labels that are strict on Spatial's referring expressions, so augmentation's insensitivity to real task changes is not established either way. The consensus conditions mix all members at every step and cannot be filtered per episode; on the \MedoidEquivTasksN{} Spatial tasks whose three members were all judged equivalent the medoid costs $\MedoidEquivTasks$ $\MedoidEquivTasksCI$, against $\MedoidOtherTasks$ $\MedoidOtherTasksCI$ on the other tasks. Lying inside the original interval is a consistency check, not an equivalence test, and the labels are one annotator's, who knew the hypotheses though not the item's type.

\begin{table}[h]
\centering\small
\caption{One-paraphrase cost on all evaluated paraphrases and on those judged equivalent by a human. Generated by \texttt{lcc/analysis/human\_equivalence.py}.}
\label{tab:human}
\resizebox{\columnwidth}{!}{\IfFileExists{tables/human_equiv.tex}{\begin{tabular}{lrlrl}
\toprule
Condition & $n$ & All paraphrases $\Delta$ [95\% CI] & $n$ & Human-equivalent only \\
\midrule
OpenVLA-OFT / Long (llm) & 100 & $-3.0$ $[-8.0, +2.0]$ & 56 & $-5.4$ $[-12.5, +0.0]$ \\
OpenVLA-OFT / Goal (llm) & 250 & $-10.4$ $[-14.4, -6.8]$ & 198 & $-8.6$ $[-12.6, -5.1]$ \\
OpenVLA-OFT / Object (llm) & 250 & $-1.6$ $[-4.0, +0.4]$ & 143 & $-2.1$ $[-5.6, +0.7]$ \\
OpenVLA-OFT / Spatial (llm) & 500 & $-1.0$ $[-3.0, +1.0]$ & 290 & $-0.3$ $[-2.8, +1.7]$ \\
base OpenVLA / Long (llm) & 100 & $-19.0$ $[-32.0, -6.0]$ & 56 & $-19.6$ $[-37.5, -1.8]$ \\
base OpenVLA / Goal (llm) & 250 & $-36.8$ $[-44.0, -29.6]$ & 198 & $-31.8$ $[-39.9, -23.7]$ \\
base OpenVLA / Object (llm) & 250 & $-15.6$ $[-22.4, -9.2]$ & 143 & $-15.4$ $[-23.8, -7.7]$ \\
base OpenVLA / Spatial (llm) & 250 & $-38.0$ $[-44.8, -31.2]$ & 143 & $-35.7$ $[-44.1, -27.3]$ \\
$\pi_{0.5}$ / Long (llm) & 100 & $-4.0$ $[-9.0, +0.0]$ & 56 & $-3.6$ $[-8.9, +0.0]$ \\
$\pi_{0.5}$ / Goal (llm) & 250 & $-7.2$ $[-11.2, -3.2]$ & 198 & $-7.1$ $[-11.6, -2.5]$ \\
$\pi_{0.5}$ / Object (llm) & 100 & $+1.0$ $[-2.0, +5.0]$ & 57 & $+1.8$ $[+0.0, +5.3]$ \\
$\pi_{0.5}$ / Spatial (llm) & 250 & $-1.2$ $[-3.6, +1.2]$ & 143 & $-2.1$ $[-5.6, +1.4]$ \\
base OpenVLA / Spatial (easy) & 250 & $-14.8$ $[-20.8, -8.8]$ & 201 & $-16.9$ $[-23.4, -10.9]$ \\
base OpenVLA / Spatial (hard) & 250 & $-48.0$ $[-55.2, -41.2]$ & 162 & $-48.8$ $[-57.4, -40.1]$ \\
base OpenVLA / Spatial (adversarial) & 250 & $-28.8$ $[-35.6, -22.4]$ & 161 & $-28.6$ $[-36.6, -20.5]$ \\
Aug / Spatial (llm) & 250 & $+2.0$ $[-4.0, +8.0]$ & 143 & $+4.9$ $[-2.8, +12.6]$ \\
Aug / Spatial (heldout) & 250 & $+2.0$ $[-3.6, +7.6]$ & 178 & $+1.7$ $[-4.5, +7.9]$ \\
Ctrl / Spatial (llm) & 250 & $-32.0$ $[-38.8, -25.2]$ & 143 & $-25.9$ $[-35.0, -17.5]$ \\
base OpenVLA / Spatial, paraphrase medoid vs.\ canonical, tasks whose members are all equivalent & 50 & $-28.0$ $[-42.0, -14.0]$ & 50 & -- \\
base OpenVLA / Spatial, paraphrase medoid vs.\ canonical, other tasks & 200 & $-14.0$ $[-21.0, -7.0]$ & 200 & -- \\
\bottomrule
\end{tabular}
}{}}
\end{table}

\IfFileExists{figs/forest_2x2.pdf}{%
\begin{widefigure}
\centering
\includegraphics[width=\figonewidth]{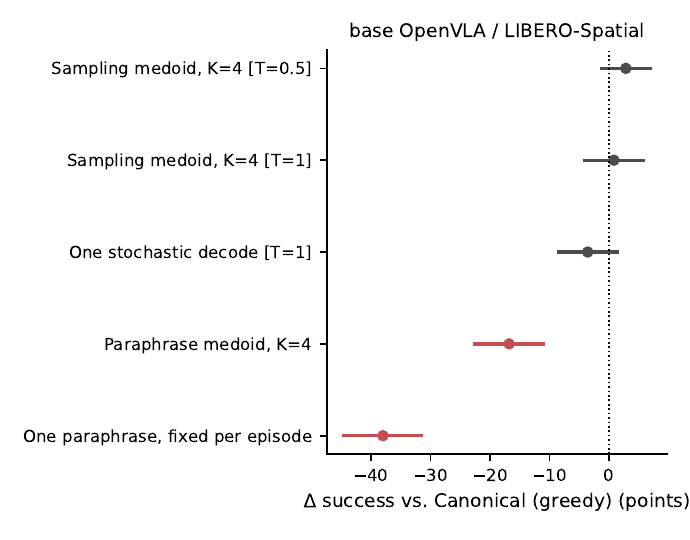}
\caption{Aggregation with and without exchangeable members, base OpenVLA / LIBERO-Spatial, paired $\Delta$ vs.\ canonical greedy with $95\%$ bootstrap intervals, $n{=}250$. Paraphrase rows: one paraphrase or the medoid of four. Sampling rows: one or four stochastic decodes of the training instruction, identical medoid code. Generated by \texttt{lcc/analysis/main\_table.py}.}
\label{fig:2x2}
\end{widefigure}}{%
\figbox{1.0cm}{\textbf{Fig.~1.} \{paraphrase, sampling\}$\times$\{single, medoid of $K{=}4\}$ on base OpenVLA / LIBERO-Spatial, paired $\Delta$ vs.\ canonical greedy with $95\%$ CIs. (figure file missing from build)}\label{fig:2x2}}

\section{What Predicts a Paraphrase's Cost}
\label{app:parafeat}
For every paraphrase an evaluation run executed, its cost is the mean success of the episodes that executed it minus the canonical success on the same episodes (about eight episodes per paraphrase, so each point is noisy). Tab.~\ref{tab:parafeat} correlates that cost with the paraphrase's overlap with the training instruction (character trigrams, word Jaccard), the pipeline's embedding cosine, the length ratio, and the human equivalence label of App.~\ref{app:human}, pooled over the suites and types each policy was evaluated on. A positive $\rho$ means more similar paraphrases cost less.

\begin{table}[h]
\centering\small
\caption{Spearman correlation of per-paraphrase cost with paraphrase features (permutation $p$ in brackets). Generated by \texttt{lcc/analysis/paraphrase\_features.py}.}
\label{tab:parafeat}
\resizebox{\columnwidth}{!}{\IfFileExists{tables/para_features.tex}{\begin{tabular}{lrrr}
\toprule
Feature (vs.\ training instruction) & base OpenVLA ($n{=}210$) & OpenVLA-OFT ($n{=}120$) & $\pi_{0.5}$ ($n{=}120$) \\
\midrule
char.-trigram cosine & $+0.36$ \tiny[$<$0.001] & $+0.21$ \tiny[0.022] & $+0.02$ \tiny[0.810] \\
word Jaccard & $+0.34$ \tiny[$<$0.001] & $+0.34$ \tiny[$<$0.001] & $+0.09$ \tiny[0.312] \\
embedding cosine & $+0.47$ \tiny[$<$0.001] & $+0.21$ \tiny[0.025] & $-0.02$ \tiny[0.801] \\
length ratio & $-0.13$ \tiny[0.071] & $-0.13$ \tiny[0.163] & $-0.20$ \tiny[0.028] \\
human: same task & $-0.00$ \tiny[0.986] & $+0.03$ \tiny[0.713] & $+0.02$ \tiny[0.836] \\
\bottomrule
\end{tabular}
}{}}
\end{table}

\section{Task-Clustered Intervals}

\label{app:clustered}
The main-text intervals resample episodes and so treat the ten tasks of a suite, and the few phrasings each task has, as fixed. Tab.~\ref{tab:clustered} resamples tasks with replacement and then episodes within each drawn task, keeping both arms of an episode together; since every phrasing belongs to one task, this also resamples phrasings.

\begin{table}[h]
\centering\small
\caption{Headline paired comparisons with task-clustered $95\%$ bootstrap intervals ($10{,}000$ resamples). Generated by \texttt{lcc/analysis/clustered\_ci.py}.}
\label{tab:clustered}
\resizebox{\columnwidth}{!}{\IfFileExists{tables/clustered.tex}{\begin{tabular}{lrrl}
\toprule
Comparison & $n$ & $\Delta$ & Task-clustered 95\% CI \\
\midrule
base OpenVLA, one paraphrase vs.\ canonical (Spatial) & 250 & $-38.0$ & $[-53.2, -21.6]$ \\
base OpenVLA, paraphrase medoid vs.\ canonical (Spatial) & 250 & $-16.8$ & $[-30.4, -4.0]$ \\
sampling medoid vs.\ paraphrase medoid (Spatial) & 250 & $+17.6$ & $[+7.2, +28.4]$ \\
base OpenVLA, paraphrase-only medoid vs.\ one paraphrase (Spatial) & 250 & $+5.6$ & $[-4.0, +16.0]$ \\
base OpenVLA, sampling medoid around a paraphrase vs.\ greedy paraphrase (Spatial) & 250 & $+10.4$ & $[+3.2, +18.0]$ \\
base OpenVLA, sampling medoid vs.\ one sample around a paraphrase (Spatial) & 250 & $+6.4$ & $[-1.6, +14.0]$ \\
OpenVLA-OFT, medoid vs.\ canonical (Goal) & 250 & $-4.0$ & $[-8.4, -0.8]$ \\
OpenVLA-OFT, paraphrase-only medoid vs.\ one paraphrase (Goal) & 250 & $+3.6$ & $[-1.2, +8.4]$ \\
augmented, one paraphrase vs.\ canonical & 250 & $+2.0$ & $[-6.0, +10.0]$ \\
augmented, held-out vs.\ canonical & 250 & $+2.0$ & $[-5.2, +8.8]$ \\
control, one paraphrase vs.\ canonical & 250 & $-32.0$ & $[-44.4, -20.0]$ \\
OpenVLA-OFT, one paraphrase vs.\ canonical (Goal) & 250 & $-10.4$ & $[-20.0, -3.6]$ \\
$\pi_{0.5}$, one paraphrase vs.\ canonical (Goal) & 250 & $-7.2$ & $[-14.4, -1.6]$ \\
\bottomrule
\end{tabular}
}{}}
\end{table}

\section{Reliance and Paraphrase Cost per Task}
\label{app:reltask}
Per task, language share and paraphrase cost are computed from the same runs as Tab.~\ref{tab:reliance} ($25$ or $10$ episodes per task, so each point is noisy). Tab.~\ref{tab:reltask} gives Spearman's $\rho$ between them, within each suite and across all tasks of a policy, with permutation $p$-values. A negative $\rho$ means higher reliance, larger cost. On Goal, and for base OpenVLA on Long, every task has a share of $100\%$ and no correlation is defined.

\begin{table}[h]
\centering\small
\caption{Spearman correlation of per-task language share with per-task paraphrase cost. Generated by \texttt{lcc/analysis/reliance\_per\_task.py}.}
\label{tab:reltask}
\resizebox{\columnwidth}{!}{\IfFileExists{tables/reliance_task.tex}{\begin{tabular}{llrrl}
\toprule
Policy & Suite & Tasks & Share range & Spearman $\rho$ (share, cost) [$p$] \\
\midrule
OpenVLA-OFT & Spatial & 10 & -14--83\% & $-0.65$ [0.04] \\
OpenVLA-OFT & Object & 10 & 0--8\% & $-0.57$ [0.10] \\
OpenVLA-OFT & Goal & 10 & 100--100\% & $--$ [nan] \\
OpenVLA-OFT & Long & 10 & -11--80\% & $-0.23$ [0.59] \\
OpenVLA-OFT & all & 40 & & $-0.56$ [0.000] \\
\midrule
base OpenVLA & Spatial & 10 & 71--100\% & $-0.52$ [0.21] \\
base OpenVLA & Object & 10 & 92--100\% & $+0.20$ [0.58] \\
base OpenVLA & Goal & 10 & 100--100\% & $--$ [nan] \\
base OpenVLA & Long & 10 & 100--100\% & $--$ [nan] \\
base OpenVLA & all & 40 & & $-0.24$ [0.139] \\
\midrule
$\pi_{0.5}$ & Spatial & 10 & 0--100\% & $-0.12$ [0.75] \\
$\pi_{0.5}$ & Object & 10 & 0--100\% & $-0.17$ [0.63] \\
$\pi_{0.5}$ & Goal & 10 & 100--100\% & $--$ [nan] \\
$\pi_{0.5}$ & Long & 10 & -50--100\% & $+0.17$ [0.66] \\
$\pi_{0.5}$ & all & 40 & & $-0.18$ [0.259] \\
\midrule
pooled & all & 120 & & $-0.48$ [0.000] \\
\bottomrule
\end{tabular}
}{}}
\end{table}

\begin{table}[h]
\centering\small
\caption{Instruction-reliance probe. Success on the original task while conditioned on another task's instruction, scene and success predicate unchanged; paired on identical initial states ($n{=}250$ per cell, $n{=}100$ on Long). Language share $=(\text{correct}-\text{wrong})/\text{correct}$. Correct-instruction cells are each probe's paired reference run. Generated by \texttt{lcc/analysis/probe\_table.py}.}
\label{tab:probe}
\resizebox{\columnwidth}{!}{\begin{tabular}{llrrr}
\toprule
Policy & Suite & Correct instr. & Wrong instr. & Language share \\
\midrule
OpenVLA-OFT & Spatial & 96.8 \tiny[93.8, 98.4] & 74.4 \tiny[68.6, 79.4] & 23\% \\
OpenVLA-OFT & Object & 98.0 \tiny[95.4, 99.1] & 96.4 \tiny[93.3, 98.1] & 2\% \\
OpenVLA-OFT & Goal & 100.0 \tiny[98.5, 100.0] & 0.0 \tiny[0.0, 1.5] & 100\% \\
OpenVLA-OFT & Long & 95.0 \tiny[88.8, 97.8] & 87.0 \tiny[79.0, 92.2] & 8\% \\
\midrule
base OpenVLA & Spatial & 85.6 \tiny[80.7, 89.4] & 2.4 \tiny[1.1, 5.1] & 97\% \\
base OpenVLA & Object & 87.6 \tiny[82.9, 91.1] & 1.2 \tiny[0.4, 3.5] & 99\% \\
base OpenVLA & Goal & 76.0 \tiny[70.3, 80.9] & 0.0 \tiny[0.0, 1.5] & 100\% \\
base OpenVLA & Long & 55.0 \tiny[45.2, 64.4] & 0.0 \tiny[0.0, 3.7] & 100\% \\
\bottomrule
\end{tabular}
}
\end{table}

\section{Is the Wrong Instruction Executable?}
\label{app:probefeas}
The probe gives task $i$'s scene the instruction of task $i{+}5$. Tab.~\ref{tab:probefeas} checks, from the benchmark's own task definitions (BDDL files, no rollouts), whether every object and fixture in task $i{+}5$'s goal exists in task $i$'s scene. On Spatial and Goal every pair is executable. On Object four pairs name a target that is absent; restricted to the six executable pairs, OFT's wrong-instruction success is \OftObjectFeasWrong\% (language share \OftObjectFeasShare\%) and base OpenVLA's is \BaseObjectFeasWrong\% (share \BaseObjectFeasShare\%), so the Object result does not depend on the absent targets. On Long no pair is executable: each long-horizon task has its own scene, and success there means the policy completed the scene's task when the instruction asked for something the scene does not contain.

\begin{table}[h]
\centering\small
\caption{Executability of the wrong instruction in the probe. Generated by \texttt{lcc/analysis/probe\_feasibility.py}.}
\label{tab:probefeas}
\resizebox{\columnwidth}{!}{\IfFileExists{tables/probe_feasibility.tex}{\begin{tabular}{lrl}
\toprule
Suite & Executable pairs & Missing goal objects (task $i\to i{+}5$) \\
\midrule
Spatial & 10/10 & -- \\
Object & 6/10 & 4$\to$9: orange\_juice; 5$\to$0: alphabet\_soup; 6$\to$1: cream\_cheese; 7$\to$2: salad\_dressing \\
Goal & 10/10 & -- \\
Long & 0/10 & 0$\to$5: black\_book; 1$\to$6: chocolate\_pudding, plate, porcelain\_mug; 2$\to$7: alphabet\_soup, basket, cream\_cheese; 3$\to$8: flat\_stove, moka\_pot; 4$\to$9: microwave; 5$\to$0: alphabet\_soup, basket, tomato\_sauce; 6$\to$1: basket, butter, cream\_cheese; 7$\to$2: flat\_stove, moka\_pot; 8$\to$3: akita\_black\_bowl, white\_cabinet; 9$\to$4: plate \\
\bottomrule
\end{tabular}
}{}}
\end{table}

\section{Closed-Set Canonicalization}
\label{app:canon}
On a closed benchmark the instruction can be mapped back to a training instruction before the query. Retrieving the nearest of a suite's ten training instructions by character trigrams returns the right one for \CanonRetTrigram\% of our \CanonRetN{} paraphrases, but for only \CanonRetTrigramWorst\% of the structural rewrites on Spatial, whose tasks differ by a referring expression and where the penalty is largest. This measures retrieval, not success, and the method presupposes the closed instruction set that deployment removes. The augmented policy needs none. A canonicalizer replaces the user's instruction by the nearest of the suite's ten training instructions and queries the policy with that. Tab.~\ref{tab:canon} reports how often the nearest training instruction is the correct task's, for every paraphrase in the released sets, with two retrievers: cosine over character-trigram counts, and cosine over the sentence embedding used by the paraphrase pipeline ($^\dagger$). The pipeline kept only candidates whose embedding cosine to their own canonical instruction cleared a floor (App.~\ref{app:paraphrases}), so the embedding retriever is favoured by construction; the trigram retriever shares nothing with that selection. No rollouts were run for this table, and retrieval accuracy is not a success rate: a mis-retrieved instruction is another task's training string, the condition of the wrong-instruction probe (\S\ref{sec:probe}).

\begin{table}[h]
\centering\small
\caption{Closed-set canonicalization: accuracy (\%) of retrieving the correct task's training instruction from the ten in the suite, by paraphrase type. Generated by \texttt{lcc/analysis/canonicalize\_retrieval.py}.}
\label{tab:canon}
\IfFileExists{tables/canon_retrieval.tex}{\begin{tabular}{@{}lrrrr@{}}
\toprule
Paraphrase type & Spatial & Object & Goal & Long \\
\midrule
\multicolumn{5}{@{}l}{\textit{character trigrams}} \\
LLM & 85.7 & 100.0 & 98.6 & 98.6 \\
lexical & 82.9 & 100.0 & 94.3 & 100.0 \\
structural & 75.7 & 100.0 & 92.9 & 98.6 \\
typos/casing & 97.1 & 100.0 & 100.0 & 100.0 \\
held-out & 88.6 & 100.0 & 98.6 & 98.6 \\
\multicolumn{5}{@{}l}{\textit{sentence embedding$^\dagger$}} \\
LLM & 75.7 & 100.0 & 100.0 & 100.0 \\
lexical & 94.3 & 100.0 & 100.0 & 100.0 \\
structural & 62.9 & 100.0 & 98.6 & 98.6 \\
typos/casing & 82.9 & 95.7 & 100.0 & 100.0 \\
held-out & 60.0 & 100.0 & 100.0 & 98.6 \\
\bottomrule
\end{tabular}
}{}
\end{table}

\section{Paraphrase Sets and Prompts}
\label{app:paraphrases}
\IfFileExists{tables/example_paraphrases.tex}{%
\begin{table}[h]
\centering\small
\caption{One LIBERO-Spatial task: the training instruction and one paraphrase of each type (App.~\ref{app:paraphrases}).}
\label{tab:example}
\begin{tabular}{@{}lp{0.74\linewidth}@{}}
\toprule
Type & Instruction \\
\midrule
training instruction & \textit{pick up the black bowl between the plate and the ramekin and place it on the plate} \\
LLM & \textit{the plate is where the black bowl between it and the ramekin needs to go} \\
lexical & \textit{retrieve the black bowl between the plate and the ramekin and settle it onto the plate} \\
structural & \textit{between the plate and the ramekin there is a black bowl, and its new spot is on the plate} \\
typos/casing & \textit{what i want is the black bowl between the plate and the ramekn picked up and put on the plate} \\
held-out & \textit{the plate is where i want the black bowl that's wedged between the plate and the ramekin, could you do that} \\
\bottomrule
\end{tabular}

\end{table}}{}
For each of the 40 LIBERO tasks and each of five variants (\emph{llm}, \emph{easy}, \emph{hard}, \emph{adversarial}, \emph{held-out}), a generator (Claude Opus~5, high effort) produced 10 candidates from the prompts below, instructed to preserve every object, attribute, spatial relation and goal. Candidates passed two filters: an independent judge on a different model (Claude Sonnet~5, medium effort) that sees only the canonical instruction and the candidate, and a sentence-embedding cosine floor of $\theta_{\text{eq}}{=}0.55$ under \texttt{all-mpnet-base-v2}. Acceptance rates were $93.2\%$ (llm), $94.0\%$ (easy), $91.0\%$ (hard), $97.9\%$ (adversarial) and $94.2\%$ (held-out); of the 118 rejections, 115 were by the judge and 3 by the embedding floor: the judge caught changes an embedding cannot see (``\emph{drop} changes manner/goal from careful placement to dropping'', cosine $0.93$). From the accepted set, 7 phrasings per variant were chosen by greedy max--min selection on embedding distance. Mean cosine to the canonical instruction is $0.95$ (easy), $0.91$ (llm), $0.91$ (adversarial), $0.86$ (hard) and $0.81$ (held-out), with mean pairwise cosine distance among the selected 7 of $0.07$, $0.15$, $0.17$, $0.19$ and $0.20$ respectively; the held-out set is the most distant from the training phrasing and the most diverse. Generation cost \$3.22 across all tasks; judging brought the corpus to \$7.25. Every paraphrase file records generator and judge models, prompt-bundle hash, per-candidate judge verdicts and cosines. The \emph{held-out} set is used only for the transfer evaluation of \S\ref{sec:aug} and appears in no training run; the \emph{adversarial} set is used only for evaluation (the type breakdown of \S\ref{sec:aug}) and appears in no training run. The prompts below are emitted verbatim from the code that used them.

\paragraph{Generation: system prompt}
\begin{Verbatim}[breaklines=true,breakanywhere=true,fontsize=\footnotesize]
You rewrite instructions given to a household robot arm. You are building an evaluation set for a robotics paper, so precision matters more than fluency.

Rules, in priority order:
1. PRESERVE MEANING EXACTLY. Every object, colour, material, spatial relation, container, and goal state in the original must survive in the rewrite, and nothing may be added. Do not specify a gripper, a hand, a speed, a trajectory, an order of sub-steps, or any detail the original leaves open.
2. DO NOT DISAMBIGUATE. If the original is vague ("the bowl"), the rewrite is equally vague. Resolving an ambiguity changes the task.
3. VARY THE SURFACE FORM. Change lexical choice and syntax: different verbs, different clause order, active/passive, imperative/declarative, relative clauses instead of prepositional chains.
4. SOUND LIKE A PERSON TALKING TO A ROBOT. Short, natural, no markup, no quotation marks, no trailing period, lower case, one clause per idea.

Return the rewrites only, through the required output format.
\end{Verbatim}

\paragraph{Generation: user prompt template}
\begin{Verbatim}[breaklines=true,breakanywhere=true,fontsize=\footnotesize]
Canonical instruction: {canonical}

Produce {n} rewrites of that instruction under the variant brief below.

Variant: {variant}
Brief: {brief}

Each rewrite must be a distinct surface form -- two rewrites that differ only by a comma count as one and are wasted. Order them from closest to the original to furthest.
\end{Verbatim}

\paragraph{Variant brief: \texttt{llm}}
\begin{Verbatim}[breaklines=true,breakanywhere=true,fontsize=\footnotesize]
Natural rephrasings of the kind different people would produce for the same request. Mix synonym substitution and syntactic restructuring freely.
\end{Verbatim}

\paragraph{Variant brief: \texttt{easy}}
\begin{Verbatim}[breaklines=true,breakanywhere=true,fontsize=\footnotesize]
Minimal edits only: substitute synonyms for the verb and for non-identifying nouns. Keep the clause structure and word order of the original.
\end{Verbatim}

\paragraph{Variant brief: \texttt{hard}}
\begin{Verbatim}[breaklines=true,breakanywhere=true,fontsize=\footnotesize]
Structural rewrites. Change the syntactic frame entirely -- e.g. turn a prepositional chain into a relative clause, front the goal instead of the object, or split one clause into two. Lexical overlap with the original should be as low as meaning preservation allows.
\end{Verbatim}

\paragraph{Variant brief: \texttt{adversarial}}
\begin{Verbatim}[breaklines=true,breakanywhere=true,fontsize=\footnotesize]
Realistic transcription and typing noise on top of the original wording: a single-character typo, a doubled letter, a missing space, inconsistent casing, or a dropped article. The words must remain recognisable -- this models a speech recogniser or a hurried typist, not an attack.
\end{Verbatim}

\paragraph{Variant brief: \texttt{heldout}}
\begin{Verbatim}[breaklines=true,breakanywhere=true,fontsize=\footnotesize]
Natural rephrasings, as in the default variant, but drawn from a different register: how someone might phrase the request more conversationally, including a polite frame or a filler opener if that is natural.
\end{Verbatim}

\paragraph{Judge: system prompt}
\begin{Verbatim}[breaklines=true,breakanywhere=true,fontsize=\footnotesize]
You audit candidate paraphrases of instructions for a household robot arm. A candidate is EQUIVALENT only if a competent robot following it would be required to do exactly the same thing as one following the original, in every situation.

Reject a candidate if it:
- drops, adds, or changes any object, colour, material, container, or spatial relation;
- resolves an ambiguity the original leaves open, or introduces an ambiguity the original resolves;
- adds a constraint the original does not impose (which hand, what order, how fast, how gently);
- changes the goal state, or turns a goal into a manner ("place it on the plate" vs "slide it toward the plate");
- is not a rewrite at all -- identical to the original, or a different task.

Accept purely surface differences: word choice, syntax, clause order, politeness, and -- for the adversarial variant only -- typos and casing that leave the words recognisable.

Be strict. A contaminated evaluation set is worse than a small one.
\end{Verbatim}

\paragraph{Judge: user prompt template}
\begin{Verbatim}[breaklines=true,breakanywhere=true,fontsize=\footnotesize]
Original instruction: {canonical}
Candidate paraphrase: {candidate}
Variant being audited: {variant}

Is the candidate equivalent to the original under the rules? If it is not, name the single most important difference in one short phrase; if it is, leave the difference empty.
\end{Verbatim}

\IfFileExists{tables/per_task_weak.tex}{%
\section{Per-Task Results}
\label{app:pertask}
Per-task success (\%) with episode counts for the base-OpenVLA comparison of \S\ref{sec:weak}, so that a reader can see whether the paraphrase penalty concentrates in particular tasks. Generated by \texttt{lcc/analysis/per\_task.py}.

\begin{widetable}
\centering\small
\resizebox{\linewidth}{!}{\begin{tabular}{lrrr}
\toprule
Task & Canonical (greedy) & Paraphrase medoid, $K{=}4$ & One paraphrase, fixed per episode \\
\midrule
0 & 84 (25) & 92 (25) & 92 (25) \\
1 & 100 (25) & 52 (25) & 32 (25) \\
2 & 88 (25) & 56 (25) & 48 (25) \\
3 & 84 (25) & 84 (25) & 84 (25) \\
4 & 72 (25) & 76 (25) & 16 (25) \\
5 & 88 (25) & 72 (25) & 44 (25) \\
6 & 96 (25) & 80 (25) & 52 (25) \\
7 & 76 (25) & 76 (25) & 28 (25) \\
8 & 88 (25) & 64 (25) & 60 (25) \\
9 & 80 (25) & 36 (25) & 20 (25) \\
\midrule
all & 85.6 (250) & 68.8 (250) & 47.6 (250) \\
\bottomrule
\end{tabular}
}
\caption{Per-task success on LIBERO-Spatial, base OpenVLA, paired on identical initial states.}
\label{tab:pertask}
\end{widetable}
}{}

\section{Statistical Details}
\label{app:stats}
\paragraph{Random-draw conditions and their paraphrase-only re-scoring.} The controller behind the rows labelled ``one of $K{=}4$ instructions'' drew uniformly from all $K$ instructions, the canonical one included, so a quarter of its episodes executed the training phrasing; the logged draw index lets those episodes be removed and the remainder paired with the canonical run on the same initial states (Tab.~\ref{tab:paraonly}). The main-text ``one paraphrase'' rows for the untouched and augmented checkpoints are instead direct runs of a paraphrase-only controller at full $n$; the two agree within their intervals.
\IfFileExists{tables/paraphrase_only.tex}{%
\begin{widetable}
\centering\small
\caption{Legacy random-draw conditions re-scored on their paraphrase-only episodes (all arms, both suites). ``All $K$ draws'' is the rate as originally run; ``paraphrase-only'' keeps the episodes whose draw was a paraphrase and pairs them with the canonical run on those episodes. Generated by \texttt{lcc/analysis/paraphrase\_only.py}.}
\label{tab:paraonly}
\resizebox{\linewidth}{!}{\begin{tabular}{lrrrrl}
\toprule
Arm & $n_{\text{para}}$ & All $K$ draws & Canonical-drawn & Paraphrase-only & $\Delta$ vs.\ canonical [95\% CI] \\
\midrule
oft:hf:libero\_spatial:llm & 369 & 96.4 & 96.9 & 96.2 & $-1.1$ $[-3.0, +0.8]$ \\
oft:hf:libero\_spatial\_with\_mug:llm & 369 & 87.8 & 87.0 & 88.1 & $-0.8$ $[-4.1, +2.4]$ \\
openvla:goal-aug-p100:libero\_goal:llm & 188 & 73.6 & 80.6 & 71.3 & $-4.8$ $[-11.7, +2.1]$ \\
openvla:hf:libero\_goal:llm & 188 & 52.4 & 75.8 & 44.7 & $-30.9$ $[-38.8, -22.3]$ \\
openvla:aug-lr1e4-s2k-p100:libero\_spatial:heldout & 188 & 84.4 & 85.5 & 84.0 & $+2.1$ $[-4.3, +8.5]$ \\
openvla:aug-lr1e4-s2k-p100:libero\_spatial:llm & 188 & 83.6 & 82.3 & 84.0 & $+2.1$ $[-5.3, +9.6]$ \\
openvla:aug-lr1e4-s2k-p50:libero\_spatial:llm & 188 & 78.4 & 80.6 & 77.7 & $-3.7$ $[-10.6, +2.7]$ \\
openvla:cons-l01:libero\_spatial:llm & 188 & 82.8 & 77.4 & 84.6 & $+3.2$ $[-3.2, +9.6]$ \\
openvla:cons-l1:libero\_spatial:heldout & 188 & 79.6 & 83.9 & 78.2 & $-6.9$ $[-13.8, -0.5]$ \\
openvla:cons-l1:libero\_spatial:llm & 188 & 82.4 & 80.6 & 83.0 & $-2.1$ $[-8.5, +3.7]$ \\
openvla:ctrl-lr1e4-s2k-p0:libero\_spatial:llm & 188 & 62.0 & 88.7 & 53.2 & $-28.2$ $[-36.2, -20.2]$ \\
openvla:aug-lr5e5-s2k-p100:libero\_spatial:llm & 188 & 80.0 & 83.9 & 78.7 & $-4.3$ $[-11.7, +3.2]$ \\
openvla:hf:libero\_spatial:llm & 188 & 58.4 & 85.5 & 49.5 & $-35.1$ $[-43.1, -26.6]$ \\
\bottomrule
\end{tabular}
}
\end{widetable}}{}
Every comparison is paired: both arms run the benchmark's fixed initial states in the same order, and the hash of each initial-state vector is logged so pairing is checked, not assumed. The paired $\Delta$ is the difference in success rate; its interval is a percentile bootstrap over \emph{episodes} ($10{,}000$ resamples, seed 0), which resamples pairs so that per-episode difficulty cancels. Significance is the exact two-sided McNemar test on the discordant pairs (the mid-$p$ variant is also computed); test and bootstrap are computed from the same pairing and cannot disagree about which episodes were compared. Within a table, $p$-values are Holm-corrected across the comparisons made against the reference. Where an interval excludes zero but $p_{\text{Holm}}{=}1$, the discordant count is small and the exact test is the one to believe. Minimum detectable effect follows from the observed discordance: at $n{=}500$ and $4.2\%$ discordance (21 of 500 episodes differing between two runs of the same controller), the smallest paired effect detectable at $\alpha{=}0.05$ with $80\%$ power is $2.6$ points, at or above the $2.2$--$2.8$ points of headroom Canonical-8 leaves on standard LIBERO-Spatial in its two runs. Unintended replicates of the baseline (App.~\ref{app:repro}) differ from it by $0.8$ points on OFT ($n{=}500$, discordance $0.01$) and $2.0$ on base OpenVLA ($n{=}250$, discordance $0.08$); the $4.2\%$ figure above comes from a third pair, two runs of one OFT controller on the same $500$ states, and is the one that sets the minimum detectable effect at that $n$. We report these as the noise floor for a paired $\Delta$. Success-rate intervals are Wilson $95\%$ intervals.

\section{Implementation Details}
\label{app:impl}

\paragraph{Batched $K$-instruction inference (OpenVLA-OFT).}
The reference action-prediction path is written for batch size one in three load-bearing places: the empty-token fixup builds a $(1,1)$ tensor, the number of prompt tokens is a scalar used to index a single fixed slice of the hidden states for the action block, and the final reshape drops the batch axis. Since paraphrases have different token lengths, the middle one binds. We implemented and \emph{measured} three strategies rather than choosing one by argument.

\emph{Grouped (used throughout).} The vision patch embeddings are computed once at batch one and broadcast. This is where the saving is, since two $224\times224$ images contribute 512 patch tokens against roughly 25 text tokens, and the $K$ prompts are then grouped by token length, with one language-model forward per distinct length. Within a group every sequence has the same length, so there is no padding and each row's token layout is identical to the single-prompt call. At $K=2$ the result is bit-identical to $K$ sequential calls; at $K=4$ and $K=8$ the largest deviation is exactly $2^{-7}$, one bfloat16 unit in the last place, arising from batch-dependent reduction order.

\emph{Padded (rejected).} Padding all $K$ prompts to a common length allows a single forward, and we tried both naive left padding and padding placed between \texttt{<BOS>} and the prompt, the latter on the reasoning that the vision patches are spliced in after position 0 on the assumption that position 0 is \texttt{<BOS>}. That reasoning predicted the second arrangement would be the better one; it measured worse. Both variants exceed our tolerance on the relative change in $\ut$ at every $K$ we tested, while the grouped strategy passes at every $K$. Padding perturbs the rotary-embedding geometry between the patch block and the prompt in proportion to the pad count, and neither arrangement escapes that; the single-forward speedup is real but is not available at an acceptable error.

Every run therefore carries an equivalence check against $K$ sequential reference calls on a real observation from its own split, gated on the deviation in units of the action scale, on the relative change in $\ut$, and on the medoid selecting the same member; a run that fails it records \texttt{batched=False} and executes sequentially rather than executing approximate actions. One caveat that the check surfaced and that matters for \S\ref{sec:medoid}: the gap between the best and second-best member's total distance can be smaller than a bfloat16 perturbation can move it, in which case \emph{which} member is the medoid is not determined at this precision: the sequential path itself selects differently between runs when the GPU's load changes kernel selection. The check therefore treats a flip as ambiguous rather than as a failure whenever the margin falls below $(K-1)\sqrt{D}\,\epsilon$ for a per-element normalized deviation $\epsilon$. This is not a defect of the implementation but a restatement of the method's own boundary: the bound is loosest exactly when the paraphrases agree, which is the regime in which \S\ref{sec:main} already shows the medoid to be within noise of the canonical controller. It is also an additional argument for breaking ties toward the canonical instruction (\S\ref{sec:medoid}).

\paragraph{Latency measurement protocol.}
Per-replan latency is wall-clock with the CUDA stream synchronised before and after. Without the synchronisation a GPU call returns when its kernels are queued and the ``latency'' is launch overhead, an order of magnitude too fast. Taking the median of 20 timed repetitions after one discarded warm-up call. Every number in Tab.~\ref{tab:cost} was taken with the card to itself, and the measurement script records the number of concurrent compute processes it observed so that a contaminated measurement is identifiable rather than silently published. This matters: an earlier attempt on a card shared with the reproduction runs returned speedups spanning $0.99\times$ to $1.69\times$ across repetitions of an \emph{identical} configuration, a spread wider than the effect being measured.

\paragraph{Hyperparameters.}
Chunk horizon $H{=}8$ for OpenVLA-OFT and $1$ for base OpenVLA, executed in full ($h{=}H$) by every fixed-horizon controller. $K{=}4$ (canonical plus three from the \emph{llm} set, chosen by max--min); $K{=}2$ and $8$ in App.~\ref{app:horizon}. Action scale $s$ is the checkpoint's own $q_{99}{-}q_{01}$ per dimension; chunk weights uniform; medoid ties break to the canonical member. Adaptive horizon: hysteresis $m{=}2$, escalation after $T{=}3$ consecutive exceedances, $\tau_{\text{low}}$ the calibration median, $\tau_{\text{high}}$ the conformal quantile at $\alpha{=}0.1$. Sampling controllers: temperature $1.0$ (and $0.5$), top-$k$ and top-$p$ disabled, decoding restricted to the 256 action tokens. Seed $7$ throughout; sampling seeds derived per (task, episode, replan). Evaluation follows the reference scripts: 10 no-op warm-up steps, step limits 220/280/300/520 for Spatial/Object/Goal/Long, $256{\times}256$ renders rotated $180^\circ$, resized to 224, centre-cropped.

\section{Post-Training Details}
\label{app:post}
LoRA rank 32 ($\alpha{=}16$, dropout 0) on the linear layers of base OpenVLA's per-suite checkpoint, using the authors' fine-tuning script unmodified except for the instruction substitution, a wrapper around the RLDS batch transform. Data: the suite's no-noops RLDS demonstrations (Spatial: 432 trajectories, ${\sim}53$k transitions), ground-truth actions, image augmentation on, shuffle buffer $8{,}000$. Each sample's instruction is replaced with probability $p$ by a paraphrase of the same task drawn uniformly from the \emph{llm}, \emph{easy} and \emph{hard} sets (210 phrasings over 10 tasks); the \emph{held-out} set is never served. The substitution rate is asserted at the end of every run and a run whose lookup fired on fewer than $90\%$ of the samples it should have is refused. Recipes (Tab.~\ref{tab:augsweep}): the main text uses learning rate $10^{-4}$ for $2{,}000$ steps with $p{=}1$ and its $p{=}0$ control; the sweep also covers $p{=}0.5$ at that rate, $5{\times}10^{-5}$ at $p{=}1$, and the first attempt at $5{\times}10^{-4}$ for $5{,}000$ steps with its own control. Batch 8 on one A100-80GB (the latencies of App.~\ref{app:cost} were measured on a 40\,GB card): $0.70$\,s per step and $36$\,GiB peak, so $5{,}000$ steps is one GPU-hour; merging the adapter into the base weights and writing the checkpoint adds about 14 minutes. The result is evaluated as an ordinary single-query policy through the same harness as every other arm.

\begin{widetable}
\centering\small
\caption{Every post-training recipe tried, base OpenVLA / LIBERO-Spatial, $n{=}250$ paired. Held-out and wrong-instruction conditions were evaluated for a subset of recipes; -- marks the rest. Paraphrase cells are one paraphrase per episode, never the training instruction, except those marked $^\ddagger$: the first recipe and its control were evaluated only with the earlier one-of-$K$ draw, which includes the training instruction in a quarter of episodes and so understates the cost. Generated by \texttt{lcc/distill/report\_p5.py}.}
\label{tab:augsweep}
\resizebox{\linewidth}{!}{\begin{tabular}{lrrrrl}
\toprule
Arm & Canonical & Paraphrase & Held-out par. & Wrong instr. & Paraphrase cost [95\% CI] \\
\midrule
untouched checkpoint & 85.6 & 47.6 & -- & 2.4 & \textbf{$-38.0$ $[-44.8, -31.2]$} \\
lr 1e-4, 2k steps, p=1 & 81.6 & 83.6 & 83.6 & 2.8 & $+2.0$ $[-4.0, +8.0]$ \\
lr 1e-4, 2k steps, p=0.5 & 82.4 & 78.4 & -- & -- & $-4.0$ $[-10.0, +2.0]$ \\
lr 5e-5, 2k steps, p=1 & 83.6 & 82.8 & -- & -- & $-0.8$ $[-6.8, +5.2]$ \\
lr 5e-4, 5k steps, p=1 & 75.6 & 74.8$^\ddagger$ & 73.6$^\ddagger$ & 5.2 & $-0.8$ $[-6.4, +4.8]$$^\ddagger$ \\
control, lr 1e-4, 2k steps & 82.8 & 50.8 & -- & -- & \textbf{$-32.0$ $[-38.8, -25.2]$} \\
control, lr 5e-4, 5k steps & 76.8 & 48.4$^\ddagger$ & 52.4$^\ddagger$ & 2.4 & \textbf{$-28.4$ $[-34.8, -21.6]$}$^\ddagger$ \\
\bottomrule
\end{tabular}
}
\end{widetable}

\begin{widetable}
\centering\small
\caption{Second training seed of the augmentation arm and its matched control (same recipe, seed 2), base OpenVLA / LIBERO-Spatial, $n{=}250$ paired per cell, beside seed 1 and the untouched checkpoint. Paired arm against arm, seed 2's augmentation beats its control by $\SeedBAugVsCtrlParaDelta$ $\SeedBAugVsCtrlParaCI$ on trained and $\SeedBAugVsCtrlHeldDelta$ $\SeedBAugVsCtrlHeldCI$ on held-out phrasings, and its paraphrase success differs from seed 1's by $\SeedBVsAAugParaDelta$ $\SeedBVsAAugParaCI$. Generated by \texttt{lcc/distill/report\_p5.py} and \texttt{lcc/analysis/arm\_compare.py}.}
\label{tab:augseed}
\resizebox{\linewidth}{!}{\begin{tabular}{lrrrrl}
\toprule
Arm & Canonical & Paraphrase & Held-out par. & Wrong instr. & Paraphrase cost [95\% CI] \\
\midrule
untouched checkpoint & 85.6 & 47.6 & 57.6 & 2.4 & \textbf{$-38.0$ $[-44.8, -31.2]$} \\
augmentation, seed 1 & 81.6 & 83.6 & 83.6 & 2.8 & $+2.0$ $[-4.0, +8.0]$ \\
augmentation, seed 2 & 83.2 & 81.2 & 84.0 & -- & $-2.0$ $[-8.0, +4.0]$ \\
control, seed 1 & 82.8 & 50.8 & 56.4 & -- & \textbf{$-32.0$ $[-38.8, -25.2]$} \\
control, seed 2 & 80.0 & 43.2 & 52.4 & -- & \textbf{$-36.8$ $[-44.0, -29.6]$} \\
\bottomrule
\end{tabular}
}
\end{widetable}

\IfFileExists{tables/reliance_vs_cost.tex}{%
\begin{widetable}
\centering\small
\caption{Language share (correct minus wrong-instruction success, over correct, with a $95\%$ percentile bootstrap interval over paired episodes) and the paraphrase-only penalty (one paraphrase per episode, paired with the canonical run) for each published checkpoint on all four suites ($\pi_{0.5}$: openpi's LIBERO client, $10$ episodes per task on Object and Long). Correct, wrong and share are computed on the probe's episodes, the paraphrase effect on all episodes paired with the canonical run. Base OpenVLA / Long's canonical run here is the probe's reference ($\BaseLongCanonProbeRef\%$); the consensus result of \S\ref{sec:weak} is paired with a separate run ($\BaseLongCanonConsRef\%$), the replicate gap of App.~\ref{app:repro}. Lost is the paraphrase effect as a fraction of the success the instruction carries, $-\Delta/(\text{correct}-\text{wrong})$, with a $95\%$ bootstrap interval, left blank where the instruction carries under $10$ points. On Long no wrong instruction is executable in the given scene (App.~\ref{app:probefeas}), so its share measures success when the instruction cannot be followed. Generated by \texttt{lcc/analysis/reliance\_vs\_cost.py}.}
\label{tab:reliance}
\begin{tabular}{llrrllrr}
\toprule
Policy & Suite & Correct & Wrong instr. & Language share [95\% CI] & One paraphrase: $\Delta$ [95\% CI] & $p_{\text{Holm}}$ & Lost \\
\midrule
OpenVLA-OFT & Spatial & 96.8 & 74.4 & 23\% \tiny[17, 29] & $-1.0$ $[-3.0, +1.0]$ & 1.000 & 4\% \tiny[-5, 15] \\
OpenVLA-OFT & Object & 98.0 & 96.4 & 2\% \tiny[0, 3] & $-1.6$ $[-4.0, +0.4]$ & 1.000 & -- \\
OpenVLA-OFT & Goal & 100.0 & 0.0 & 100\% \tiny[100, 100] & \textbf{$-10.4$ $[-14.4, -6.8]$} & $<$0.001 & 10\% \tiny[7, 14] \\
OpenVLA-OFT & Long & 95.0 & 87.0 & 8\% \tiny[2, 16] & $-3.0$ $[-8.0, +2.0]$ & 1.000 & -- \\
\midrule
base OpenVLA & Spatial & 85.6 & 2.4 & 97\% \tiny[95, 99] & \textbf{$-38.0$ $[-44.8, -31.2]$} & $<$0.001 & 46\% \tiny[37, 55] \\
base OpenVLA & Object & 87.6 & 1.2 & 99\% \tiny[97, 100] & \textbf{$-15.6$ $[-22.4, -9.2]$} & $<$0.001 & 18\% \tiny[11, 26] \\
base OpenVLA & Goal & 76.0 & 0.0 & 100\% \tiny[100, 100] & \textbf{$-36.8$ $[-44.0, -29.6]$} & $<$0.001 & 48\% \tiny[39, 59] \\
base OpenVLA & Long & 55.0 & 0.0 & 100\% \tiny[100, 100] & $-19.0$ $[-32.0, -6.0]$ & 0.066 & 35\% \tiny[10, 62] \\
\midrule
$\pi_{0.5}$ & Spatial & 98.8 & 72.4 & 27\% \tiny[21, 32] & $-1.2$ $[-3.6, +1.2]$ & 1.000 & 5\% \tiny[-4, 14] \\
$\pi_{0.5}$ & Object & 98.0 & 32.0 & 67\% \tiny[58, 77] & $+1.0$ $[-2.0, +5.0]$ & 1.000 & -2\% \tiny[-7, 3] \\
$\pi_{0.5}$ & Goal & 97.2 & 0.0 & 100\% \tiny[100, 100] & \textbf{$-7.2$ $[-11.2, -3.2]$} & 0.007 & 7\% \tiny[3, 12] \\
$\pi_{0.5}$ & Long & 95.0 & 58.0 & 39\% \tiny[28, 50] & $-4.0$ $[-9.0, +0.0]$ & 1.000 & 11\% \tiny[-0, 26] \\
\bottomrule
\end{tabular}

\end{widetable}}{}

\IfFileExists{tables/variants.tex}{%
\begin{widetable}
\centering\small
\caption{Paraphrase penalty by paraphrase type, base OpenVLA / LIBERO-Spatial: one paraphrase drawn per episode from the $K{-}1{=}3$ paraphrases of the given type (never the canonical instruction), paired with each checkpoint's own canonical run. The augmented checkpoint trained on the LLM, lexical and structural sets only; typos/casing and the register-shifted held-out set were never seen. Generated by \texttt{lcc/analysis/variant\_table.py}.}
\label{tab:variants}
\begin{tabular}{llrrlr}
\toprule
Policy & Paraphrase type & $n$ & Success (\%) & $\Delta$ vs.\ canonical [95\% CI] & $p_{\text{Holm}}$ \\
\midrule
augmented (p=1) & LLM paraphrases (default) & 250 & 83.6 \tiny[78.5, 87.7] & $+2.0$ $[-4.0, +8.0]$ & 1.000 \\
augmented (p=1) & lexical substitution (easy) & 250 & 82.4 \tiny[77.2, 86.6] & $+0.8$ $[-4.8, +6.4]$ & 1.000 \\
augmented (p=1) & structural rewrite (hard) & 250 & 79.2 \tiny[73.7, 83.8] & $-2.4$ $[-8.4, +3.6]$ & 1.000 \\
augmented (p=1) & typos and casing (adversarial) & 250 & 74.8 \tiny[69.1, 79.8] & $-6.8$ $[-13.2, -0.4]$ & 0.249 \\
augmented (p=1) & register shift (held-out) & 250 & 83.6 \tiny[78.5, 87.7] & $+2.0$ $[-3.6, +7.6]$ & 1.000 \\
\midrule
base OpenVLA & LLM paraphrases (default) & 250 & 47.6 \tiny[41.5, 53.8] & \textbf{$-38.0$ $[-44.8, -31.2]$} & $<$0.001 \\
base OpenVLA & lexical substitution (easy) & 250 & 70.8 \tiny[64.9, 76.1] & \textbf{$-14.8$ $[-20.8, -8.8]$} & $<$0.001 \\
base OpenVLA & structural rewrite (hard) & 250 & 37.6 \tiny[31.8, 43.7] & \textbf{$-48.0$ $[-55.2, -41.2]$} & $<$0.001 \\
base OpenVLA & typos and casing (adversarial) & 250 & 56.8 \tiny[50.6, 62.8] & \textbf{$-28.8$ $[-35.6, -22.4]$} & $<$0.001 \\
\bottomrule
\end{tabular}

\end{widetable}}{}

\paragraph{The explicit consistency term, in full.} Augmentation reduces the disagreement of \S\ref{sec:method} only indirectly; we also trained it directly: a second forward on the canonical instruction with the same pixels, and $\lambda$ times the symmetric KL between the two action-token distributions at the seven action positions, the behaviour-cloning term split equally ($0.5\,\mathrm{CE}_{\text{para}}+0.5\,\mathrm{CE}_{\text{canon}}$). The matched control is the $p{=}0.5$ mix without the KL. At step one the untouched policy's cross-entropy is \ConsFirstCECanon{} on the canonical instruction and \ConsFirstCEPara{} on a paraphrase, \ConsFirstKL{} nats of KL per action position apart: the privilege of the training instruction, read off the loss. With $\lambda{=}1$ the post-trained policy reaches \ConsACanon\% canonical, \ConsAPara\% under paraphrase (cost $\ConsACost$ $\ConsACostCI$), \ConsAHeldout\% on held-out phrasings, and a language share of \ConsAShare\%, against the $p{=}0.5$ control's paraphrase cost of $\HalfCost$ $\HalfCostCI$; $\lambda{=}0.1$ gives $\ConsBCost$ $\ConsBCostCI$ (App.~\ref{app:post}\IfFileExists{tables/aug_cons.tex}{, Tab.~\ref{tab:augcons}}{}). Relative to augmentation alone ($p{=}1$), paired on the same states, the explicit term changes canonical success by $\ConsVsAugCanonDelta$ $\ConsVsAugCanonCI$, paraphrase success by $\ConsVsAugParaDelta$ $\ConsVsAugParaCI$ and held-out success by $\ConsVsAugHeldoutDelta$ $\ConsVsAugHeldoutCI$: nothing distinguishable from zero, with transfer and the wrong-instruction rate (\ConsAWrong\%) both trending the wrong way. The KL it minimises ends at \ConsLastKL{} nats; augmentation was already driving the two conditionings together without being told to.

\IfFileExists{tables/aug_cons.tex}{%
\begin{widetable}
\centering\small
\caption{Explicit paraphrase-consistency term against augmentation alone, base OpenVLA / LIBERO-Spatial, $n{=}250$ paired. All arms share learning rate $10^{-4}$ and $2{,}000$ steps. Generated by \texttt{lcc/distill/report\_p5.py}.}
\label{tab:augcons}
\resizebox{\linewidth}{!}{\begin{tabular}{lrrrrl}
\toprule
Arm & Canonical & Paraphrase & Held-out par. & Wrong instr. & Paraphrase cost [95\% CI] \\
\midrule
untouched checkpoint & 85.6 & 47.6 & -- & 2.4 & \textbf{$-38.0$ $[-44.8, -31.2]$} \\
augmentation, p=1 & 81.6 & 83.6 & 83.6 & 2.8 & $+2.0$ $[-4.0, +8.0]$ \\
augmentation, p=0.5 (matched control) & 82.4 & 78.4 & -- & -- & $-4.0$ $[-10.0, +2.0]$ \\
+ consistency, lambda=1 & 84.4 & 81.6 & 81.2 & 4.8 & $-2.8$ $[-8.0, +2.4]$ \\
+ consistency, lambda=0.1 & 81.2 & 82.4 & -- & -- & $+1.2$ $[-4.4, +7.2]$ \\
\bottomrule
\end{tabular}
}
\end{widetable}}{}

\IfFileExists{tables/aug_consensus.tex}{%
\begin{widetable}
\centering\small
\caption{Paraphrase consensus on the augmented checkpoint, paired against that checkpoint's own canonical run. Generated by \texttt{lcc/analysis/main\_table.py}.}
\label{tab:augconsensus}
\begin{tabular}{lrrlr}
\toprule
Method & $n$ & Succ.\ (\%) & $\Delta$ vs.\ ref.\ [95\% CI] & $p_{\text{Holm}}$ \\
\midrule
\multicolumn{5}{l}{\textit{base OpenVLA / LIBERO-Spatial, reference: Canonical (greedy)}} \\
\midrule
Canonical (greedy) & 250 & 81.6 \tiny[76.3, 85.9] & --- & --- \\
Paraphrase medoid, $K{=}4$ & 250 & 84.8 \tiny[79.8, 88.7] & +3.2 [-2.4, 8.8] & 0.322 \\
\bottomrule
\end{tabular}

\end{widetable}}{}

\begin{widetable}
\centering\small
\caption{Instruction augmentation on LIBERO-Goal (learning rate $10^{-4}$, $2{,}000$ steps), base OpenVLA, $n{=}250$ paired. The untouched checkpoint's canonical and wrong-instruction cells are its probe runs of \S\ref{sec:probe}; its paraphrase cell was run for this table. Generated by \texttt{lcc/distill/report\_p5.py}.}
\label{tab:auggoal}
\resizebox{\linewidth}{!}{\begin{tabular}{lrrrrl}
\toprule
Arm & Canonical & Paraphrase & Held-out par. & Wrong instr. & Paraphrase cost [95\% CI] \\
\midrule
untouched checkpoint & 76.0 & 39.2 & -- & 0.0 & \textbf{$-36.8$ $[-44.0, -29.6]$} \\
augmentation, p=1 & 78.4 & 74.4 & -- & 0.0 & $-4.0$ $[-10.0, +2.0]$ \\
canonical-only control & 78.0 & 34.4 & -- & -- & \textbf{$-43.6$ $[-50.4, -36.4]$} \\
\bottomrule
\end{tabular}
}
\end{widetable}

OpenVLA-OFT is post-trained from its released LIBERO-Goal checkpoint with OFT's own fine-tuning script, the same instruction substitution, LoRA rank 32, learning rate $10^{-4}$ and $2{,}000$ steps at an effective batch of $8$. Three things differ from the base-OpenVLA recipe: the run resumes from the released step-$50{,}000$ action head and proprioception projector so they are trained rather than re-initialised; it trains on a private copy of the teacher, because the trainer rewrites files in the model directory; and on the $40$\,GB cards available a micro-batch of $2$ with $4$ accumulation steps replaces batch $8$ ($2.4$\,s per step, $27$\,GiB peak).

\begin{widetable}
\centering\small
\caption{Instruction augmentation on OpenVLA-OFT / LIBERO-Goal, $n{=}250$ paired, against a matched control trained on the training instructions and the untouched checkpoint. Generated by \texttt{lcc/distill/report\_p5.py}.}
\label{tab:augoftgoal}
\resizebox{\linewidth}{!}{\IfFileExists{tables/aug_oft_goal.tex}{\begin{tabular}{lrrrrl}
\toprule
Arm & Canonical & Paraphrase & Held-out par. & Wrong instr. & Paraphrase cost [95\% CI] \\
\midrule
untouched checkpoint & 100.0 & 89.6 & 92.0 & 0.0 & \textbf{$-10.4$ $[-14.4, -6.8]$} \\
augmentation, p=1 & 96.8 & 96.8 & 94.8 & 0.0 & $+0.0$ $[-2.4, +2.0]$ \\
canonical-only control & 97.2 & 87.2 & 90.0 & 0.0 & \textbf{$-10.0$ $[-14.4, -5.6]$} \\
\bottomrule
\end{tabular}
}{}}
\end{widetable}

\end{document}